\documentclass[10pt,twocolumn,letterpaper]{article}
\usepackage{cvpr}              % no [review]
\usepackage[table]{xcolor}

\usepackage{graphicx}
\usepackage{booktabs}
\usepackage{makecell}
\usepackage{adjustbox}
\usepackage{xcolor}
\usepackage{multirow}

\definecolor{lightred}{RGB}{255,220,220}
\definecolor{lightorange}{RGB}{255,235,210}
\definecolor{lightyellow}{RGB}{255,245,200}

\usepackage{hyperref}

\usepackage{orcidlink}

\begin{document}

\title{ReMATF: Recurrent Motion-Adaptive Multi-scale Turbulence Mitigation for Dynamic Scenes} 


   \author{Zhiming Liu, Zhicheng Zou, and Nantheera Anantrasirichai\\
   Visual Information Laboratory, School of Computer Science, University of Bristol\\
   }

\maketitle

\begin{abstract}
Atmospheric turbulence severely degrades video quality by introducing distortions such as geometric warping, blur, and temporal flickering, posing significant challenges to both visual clarity and temporal consistency. Current state-of-the-art methods are based on 3D architectures and require multi-frame input, but their large computational cost and memory usage limit real-time deployment. Here, we propose ReMATF, a lightweight recurrent framework that restores videos using only two frames at a time while preserving spatial detail and temporal stability. ReMATF combines a multi-scale encoder–decoder with temporal warping and a motion-adaptive temporal fusion module that performs per-pixel fusion between the warped previous output and the current prediction to enhance coherence without enlarging the temporal window. This design reduces flicker, sharpens details, and remains efficient. Experiments on synthetic and real turbulence datasets show consistent improvements in PSNR($+$2.83), SSIM($+$0.05) and LPIPS(-0.048), along with substantially faster inference, making ReMATF suitable for turbulence mitigation in resource-constrained scenarios.
\end{abstract}

%  \keywords{Atmospheric turbulence \and Restoration \and Denoising}
% -------------------------------------------
\section{Introduction}
\label{sec:intro}

Atmospheric turbulence (AT) causes image degradation in long-range imaging systems due to spatial and temporal variations in the air’s refractive index along the optical path~\cite{andrews1999optical}. The primary effects include geometric distortion, where images appear to shimmer, warp, and ripple as turbulent air motion perturbs the light path. Rapid turbulent fluctuations act as a time-varying lens, smearing fine details and resulting in image blur~\cite{rana2008generic}. In addition, intensity fluctuations cause random local brightening and dimming, an effect that becomes more pronounced at longer ranges and with smaller aperture sizes. These distortions not only deteriorate visual quality but also impair scene interpretation by both humans and machines, particularly under severe conditions.

Mitigating AT distortion is therefore essential. Prior to the advent of deep learning, traditional approaches such as non-rigid registration, multi-frame fusion, and deconvolution were commonly employed \cite{5540158, Anantrasirichai:Atmospheric:2013, 6178259}. These methods can produce clean and stable results, but are largely limited to static scenes, where hundreds of frames can be exploited under the assumption that pixel displacements have approximately zero mean and exhibit quasi-periodic motion \cite{li2009suppressing, Hill2025}. Consequently, effective AT mitigation in dynamic scenes remains challenging, as temporal redundancy for the same objects is insufficient. 

Deep learning has emerged as a promising solution, as models can be trained on large collections of distorted sequences rather than relying on per-sequence processing, as in traditional image processing methods~\cite{chan2023computational,Hill2025}. However, these data-driven approaches (discussed in Sec.~\ref{sec:analysis}) are often trained on synthetic datasets due to the lack of ground-truth data for AT. As a result, they typically perform well only within the distortion regimes seen during training. Moreover, in real-world scenarios, turbulence severity can vary significantly, limiting the generalization capability of existing methods. Consistent with traditional approaches, static scenes benefit from an increased number of input frames, leading to improved restoration quality \cite{Zhang_2025_MambaTM}. However, the number of frames is constrained by system memory. Notably, the trade-off between temporal information in scenes containing both static backgrounds and moving objects remains underexplored.

This paper focuses on AT mitigation in dynamic, real-world scenes. We systematically analyze the influence of distortion severity, training data composition, and temporal window size on generalizable restoration performance. To address memory limitations while preserving long-range temporal cues, we introduce a recurrent transformer-based framework that explicitly separates short- and long-term temporal reasoning.
In summary, we offer the following contributions:
\begin{itemize}[leftmargin=*]
    \item We propose a novel lightweight method, \textbf{ReMATF}, to mitigate AT distortions through coupled short- and long-term recurrent mechanisms. The short-term recurrence suppresses ripple artifacts using restoration cues and long-term recurrence enables memory-efficient temporal aggregation over extended sequences.
% where information from earlier frames is implicitly attenuated through recurrent propagation.
    \item We introduce \textbf{Motion-Adaptive Temporal Fusion (MATF)}, which performs pixel-wise fusion between the current restoration and a motion-compensated previous output. By assigning greater reliance on the current estimate in dynamic regions and on historical information in static regions, MATF reduces ghosting while suppressing temporal flicker.
    \item We introduce turbulence-level conditioning to adapt feature processing to varying distortion severities, and a pseudo-dynamic training protocol that derives paired samples from real turbulent static scenes through simulated camera translation.
\end{itemize}
Experiments on synthetic and real turbulence datasets demonstrate improved spatial restoration ($+$2.83 dB PSNR on dynamic scenes) and temporal consistency (18\% on tLPIP), while maintaining the efficiency of a two-frame recurrent design.

%The experiments show that ...

% -------------------------------------------
\section{Related Work}
\label{sec:relatedwork}
\noindent
Early learning based video approaches primarily use residual CNN restoration with limited explicit temporal modelling. Gao et al.\ \cite{gao2019atmosphericturbulenceremovalusing} introduce a DnCNN-style network and exploit temporal cues via  multi-frame aggregation. To incorporate stronger alignment priors, Hoffmire et al.\ \cite{Hoffmire2021} apply block matching-based registration and multi-shot averaging followed by CNN refinement. Later, Anantrasirichai \cite{anan2023atmospheric} explores complex-valued convolution, motivated by the need to better model phase information altered by AT~\cite{9858990}. 
Generative and adversarial learning have also been explored. Jin et al.\ \cite{Jin2021Gan} propose TSR-WGAN that integrates spatial and temporal information in a 3D input, and employs a large-scale dataset spanning algorithm-simulated, physically simulated, and real sequences. TurbuGAN \cite{Feng2023TurbuGAN} removes the need for paired ground truth by combining a physics-based forward model with adversarial distribution matching between captured and simulated measurements for multi-frame blind deconvolution. ATVR GAN \cite{Ettedgui2023ATVR_GAN} introduces a recurrent GAN-based video restoration approach that estimates turbulent optical flow and propagates temporal information. Pyramid GAN and inverting GAN are used in \cite{YUAN2025112880} and \cite{10023498}, respectively.

Recent methods aggregate temporal information using attention mechanisms or recurrent designs. TMT \cite{Zhang2024TMT} introduces a two-stage pipeline that decouples tilt correction from deblurring and extracts features using temporal attention. PiRN \cite{Jaiswal:Physics:2023} incorporates physics-based turbulence priors into a transformer framework. DATUM \cite{Zhang2024DATUM} translates the classical lucky imaging principle into a learnable architecture, using deformable attention and temporal channel attention TurbSegRes \cite{saha2024turb} proposes a segment-then-restore pipeline, separating moving foreground from background before restoration. RMFAT \cite{Liu2025RMFAT} employs a recurrent multi-scale feature design that iteratively refines restoration across scales. DeTurb \cite{Zou:deturb:2024} combines deformable 3D convolutions for non-rigid alignment with a 3D Swin-based enhancement network. NeRT \cite{Jiang:NeRT:2023} leverages implicit neural representations together with a physically grounded tilt-then-blur turbulence model, while Nair \cite{Nair:AT-DDPM:2023} employs diffusion models.

In parallel, generic video restoration backbones have been adapted to turbulence, such as EDVR \cite{david2024EDVR} style designs that leverage deformable alignment and multi-frame fusion to improve stability. Relatedly, generic video restoration architectures originally proposed for video deblurring can also serve as transferable backbones for AT studies, including the transformer-based VRT \cite{10462902}, the efficient spatio-temporal recurrent ESTRNN \cite{zhong2023real}, and recurrent multi-scale bi-directional propagation designs such as RNN-MBP \cite{Zhu_Dong_Pan_Liang_Huang_Fu_Wang_2022}. 

To scale temporal modelling without the quadratic complexity of self-attention, MambaTM \cite{Zhang_2025_MambaTM} and MAMAT \cite{hill2025mamat} apply selective state space models for long-range spatiotemporal aggregation, and introduce a learned latent phase distortion representation to provide turbulence-specific guidance. Task-driven formulations further couple mitigation with downstream objectives, such as DMAT \cite{Hill2026DMAT}, which jointly trains an AT mitigator and an object detector in an end-to-end framework, enabling feature exchange between low-level restoration and high-level semantics. 

\newcommand{\mtriple}[3]{#1/#2/#3}
\newcommand{\trainlevel}[1]{\textcolor{gray}{\scriptsize\,(\,#1\,)}}

\begin{table*}[tb]
  \caption{Performance comparison of models trained and tested with AT levels.}
  \label{tab:difflevels}
  \centering
  \setlength{\tabcolsep}{4pt}
  \renewcommand{\arraystretch}{1.2}
  \begin{adjustbox}{width=0.85\linewidth,center}
  \begin{tabular}{@{}lcccc@{}}
    \toprule
    Model (trained on) & \multicolumn{3}{c}{Tested on} & Avg. \\
    \cmidrule(lr){2-4}
     & Weak & Medium & Strong & PSNR/SSIM/LPIPS \\
     & PSNR/SSIM/LPIPS & PSNR/SSIM/LPIPS & PSNR/SSIM/LPIPS &  \\
    \midrule

    MAMAT\cite{hill2025mamat}\trainlevel{weak}   &
    \mtriple{\textbf{27.6543}}{\textbf{0.8426}}{\textbf{0.2226}} &
    \mtriple{27.3746}{0.8207}{0.2395} &
    \mtriple{24.4744}{0.7580}{0.2955} &
    \mtriple{26.5011}{0.8071}{0.2525} \\

    MAMAT\cite{hill2025mamat}\trainlevel{medium} &
    \mtriple{27.5451}{0.8417}{0.2234} &
    \mtriple{27.4546}{0.8225}{\textbf{0.2376}} &
    \mtriple{24.5057}{0.7591}{0.2952} &
    \mtriple{26.5018}{0.8078}{0.2521} \\

    MAMAT\cite{hill2025mamat}\trainlevel{strong} &
    \mtriple{27.5166}{0.8415}{0.2290} &
    \mtriple{\textbf{27.6966}}{\textbf{0.8258}}{0.2434} &
    \textbf{\mtriple{24.8913}{0.7675}{0.2944}} &
    \textbf{\mtriple{26.7015}{0.8116}{0.2556}} \\

    \midrule

    MambaTM\cite{Zhang_2025_MambaTM}\trainlevel{weak}   &
    \mtriple{27.9953}{\textbf{0.8541}}{\textbf{0.1972}} &
    \mtriple{27.4776}{0.8401}{0.2001} &
    \mtriple{26.0382}{0.7969}{0.2391} &
    \mtriple{27.1704}{0.8304}{0.2121} \\

    MambaTM\cite{Zhang_2025_MambaTM}\trainlevel{medium} &
    \mtriple{\textbf{28.0181}}{0.8535}{0.1979} &
    \textbf{\mtriple{27.6138}{0.8422}{0.1980}} &
    \mtriple{26.2394}{0.8015}{0.2388} &
    \textbf{\mtriple{27.2904}{0.8324}{0.2116}} \\

    MambaTM\cite{Zhang_2025_MambaTM}\trainlevel{strong} &
    \mtriple{27.7253}{0.8456}{0.2079} &
    \mtriple{27.4436}{0.8366}{0.2050} &
    \textbf{\mtriple{26.2879}{0.8018}{0.2366}} &
    \mtriple{27.1523}{0.8280}{0.2165} \\

    \bottomrule
  \end{tabular}
  \end{adjustbox}
\end{table*}
\begin{table*}[tb]
  \caption{Performance comparison of models trained and tested with different datasets.}
  \label{tab:diffdatasets}
  \centering
  %\footnotesize
  \setlength{\tabcolsep}{4pt}
  \renewcommand{\arraystretch}{1.1}
  \begin{adjustbox}{width=0.7\linewidth,center}
  \begin{tabular}{@{}lccc@{}}
    \toprule
    Model (trained on) & \multicolumn{3}{c}{Tested on} \\
    \cmidrule(lr){2-4}
     & ATSyn-Dynamic & DeTurb & Both \\
     & PSNR/SSIM/LPIPS & PSNR/SSIM/LPIPS & PSNR/SSIM/LPIPS \\
    \midrule

    MAMAT\cite{hill2025mamat}\trainlevel{ATSyn-Dynamic} &
    \textbf{\mtriple{24.8407}{0.7678}{0.2852}} &
    \mtriple{26.7033}{0.7991}{0.2941} &
    \mtriple{26.3236}{0.7882}{0.2923} \\

    MAMAT\cite{hill2025mamat}\trainlevel{DeTurb} &
    \mtriple{24.3604}{0.7575}{0.2923} &
    \textbf{\mtriple{26.7887}{0.8025}{0.2808}} &
    \mtriple{26.2239}{0.7871}{0.2884} \\

    MAMAT\cite{hill2025mamat}\trainlevel{Both} &
    \mtriple{24.3088}{0.7567}{0.2978} &
    \mtriple{26.0158}{0.7843}{0.2942} &
    \textbf{\mtriple{26.5229}{0.7985}{0.2868}} \\

    \midrule

    MambaTM\cite{Zhang_2025_MambaTM}\trainlevel{ATSyn-Dynamic} &
    \textbf{\mtriple{26.3161}{0.7995}{0.2419}} &
    \mtriple{26.7766}{0.8040}{0.2447} &
    \mtriple{26.6844}{0.8033}{0.2423} \\

    MambaTM\cite{Zhang_2025_MambaTM}\trainlevel{DeTurb} &
    \mtriple{25.6761}{0.7824}{0.2508} &
    \textbf{\mtriple{27.1868}{0.8204}{0.2289}} &
    \mtriple{26.8199}{0.8109}{0.2366} \\

    MambaTM\cite{Zhang_2025_MambaTM}\trainlevel{Both} &
    \mtriple{26.1983}{0.7969}{0.2467} &
    \mtriple{27.1154}{0.8175}{0.2353} &
    \textbf{\mtriple{26.9103}{0.8127}{0.2332}} \\

    \bottomrule
  \end{tabular}
  \end{adjustbox}
\end{table*}
% ============================================
\section{Analysis of Generalisation in AT Mitigation}
\label{sec:analysis}

We employ two methods for our analysis: MAMAT \cite{hill2025mamat} and MambaTM \cite{Zhang_2025_MambaTM}. Both are based on the Mamba architecture, which has been shown to outperform alternative models in terms of both restoration quality and computational efficiency \cite{Hill2025}. Details of the datasets used are provided in \cref{sec:datasets}.

% -------------------------------------------
\vspace{2mm}
\noindent\textbf{Effect of Turbulence Severity in Training Data.}
\label{ssec:severe}
We first investigate the impact of turbulence severity in the training data. We conduct this study on the ATSyn-Dynamic dataset \cite{Zhang2024DATUM}, which provides three distortion levels: Weak, Medium, and Strong. Cross-level training and evaluation results are summarised in \cref{tab:difflevels}. Our experiments reveal that models trained on a narrow distortion range generalise poorly outside that regime. In particular, models trained only on subtle turbulence perform well on weak AT but degrade significantly when evaluated under severe conditions. 
A consistent pattern emerges across both MAMAT and MambaTM. Performance on Strong test data improves when training includes Strong turbulence, whereas models trained solely on Weak exhibit the largest degradation when evaluated on Strong. For MAMAT, the best Medium and Strong results are obtained when trained on Strong, while optimal Weak performance is achieved when trained on Weak. This demonstrates a clear trade-off between specialisation to subtle distortions and robustness to severe turbulence. MambaTM exhibits the same tendency and further suggests that the optimal training severity for Weak and Medium depends on the evaluation metric, with PSNR and perceptual metrics favouring different regimes.

Although training on mixed-severity data substantially improves robustness, it remains insufficient when the model lacks explicit knowledge of the turbulence level at inference time. To address this limitation, we introduce \textit{turbulence-level conditioning} (\cref{ssec:turbcondition}). We employ CLIP-IQA~\cite{wang2022exploring} to estimate perceptual quality and feed the resulting score as an auxiliary input to the network.

% -------------------------------------------
\vspace{2mm}
\noindent\textbf{Synthetic Dataset Diversity and Generalisation.}
\label{ssec:syndiversity}
We investigate the impact of synthetic dataset diversity by evaluating models trained on atmospheric turbulence data generated using different simulation methods. Specifically, we employ two synthetic datasets: ATSyn-Dynamic with strong AT levels \cite{Zhang2024DATUM} and the DeTurb dataset \cite{Zou:deturb:2024}, training separate models on each. \cref{tab:diffdatasets} summarises the results and reveals a clear dependence on the synthetic generation process. For both MAMAT and MambaTM, the best performance is consistently achieved when training and testing are performed on the same synthetic dataset, whereas cross-dataset evaluation leads to noticeable degradation. Training on the combined set (ATSyn plus DeTurb) improves the overall performance on the merged test set, but does not recover the in-domain peaks on either dataset, indicating that simply mixing two synthetic sources is insufficient to remove simulator-specific biases. Although MambaTM attains higher overall scores than MAMAT, it exhibits the same generalisation pattern, suggesting that the limitation is primarily data-driven rather than architecture-specific.

\begin{figure*}[tb]
    \centering
    \includegraphics[width=1.0\linewidth]{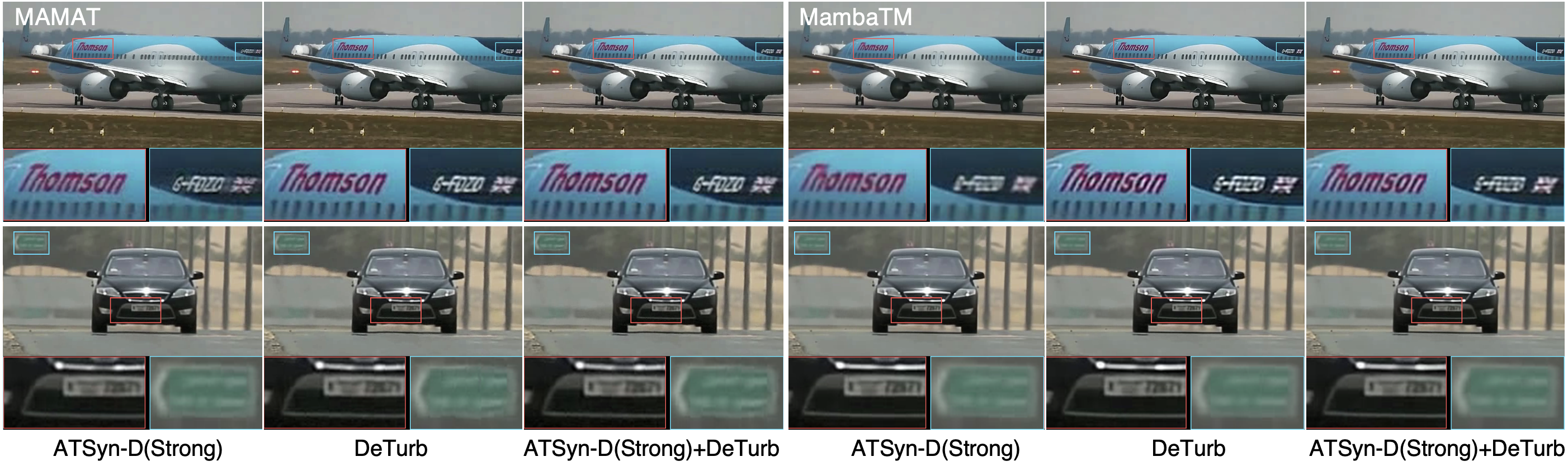}
    \caption{Restored results of real AT distortions from MAMAT \cite{hill2025mamat} and MambaAT \cite{Zhang_2025_MambaTM}, trained on different synthetic datasets.}
    \label{fig:analyseRealAT}
\end{figure*}

\cref{fig:analyseRealAT} shows the performance of models evaluated on real atmospheric turbulence footage, highlighting the impact of different synthetic training datasets despite using the same network architectures. The results indicate that training on even both synthetic datasets is insufficient to achieve robustness to real-world atmospheric turbulence. To enable the model to learn real AT distortions, we include real-world videos in the training dataset. Since corresponding clean versions are not available, we generate pseudo-ground truth using traditional image processing methods \cite{Anantrasirichai:Atmospheric:2013}, which produce desirable results for static scenes. We then simulate dynamic scenes by mimicking camera translation at various speeds.

% -------------------------------------------
\vspace{2mm}
\noindent\textbf{Influence of Temporal Window Size.}
In principle, a longer temporal window improves turbulence mitigation. In static regions, aggregating more frames suppresses ripple artefacts through temporal averaging, while in dynamic regions similar benefits can be achieved with accurate frame alignment. However, jointly processing many frames is computationally expensive and memory intensive.
To enable long-range temporal aggregation without increasing memory usage, we adopt a recurrent formulation that propagates information through the previously restored frame. The current intermediate restoration $\hat{O}_t$ is fused with the previous restored frame $O_{t-1}$ with exponentially decaying weight: $
O_t = M \hat{O}_t + (1 - M) O_{t-1},
$
where earlier information decays progressively through recursive propagation~\cite{8451755}.

\begin{figure}[tb]
    \centering
    \includegraphics[width=1.0\linewidth, trim=0 0 0 2cm, clip]{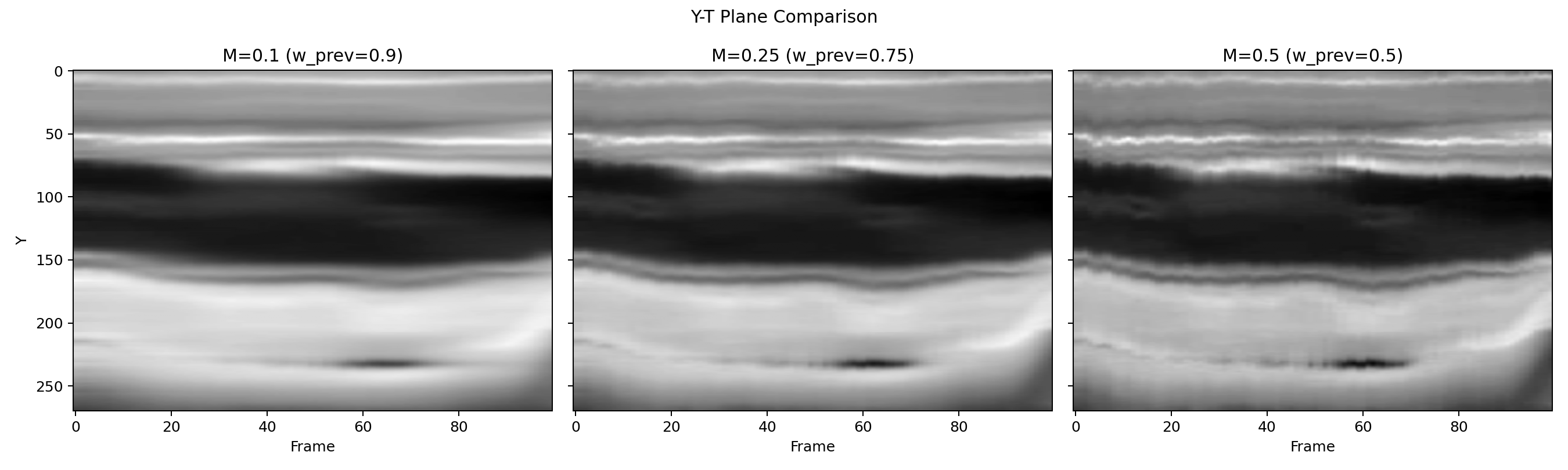}
    \caption{$y$--$t$ plane visualisation comparing recurrent temporal fusion with different weights $M\in\{0.1,0.25,0.5\}$ under severe turbulence.}
    \label{fig:yt-plane}
\end{figure}

We analyse the effect of the fusion weight by varying $M \in \{0.1, 0.25, 0.5\}$. Smaller values of $M$ place greater emphasis on the previous restoration, effectively enlarging the temporal window. As shown in \cref{fig:yt-plane}, under strong turbulence, smaller $M$ produces smoother and more stable reconstructions by leveraging accumulated information from earlier frames, which helps suppress ripple artefacts and geometric distortions. However, object motion inevitably introduces occlusions, and imperfect alignment may produce artefacts when historical information dominates. Consequently, for weaker turbulence a larger $M$ is preferable. When $M=0.5$, the current estimate and previous output contribute equally, providing sufficient smoothing while maintaining temporal responsiveness. This setting corresponds to the short-term recurrent design in our framework (see \cref{fig:architecture}), where only limited historical information is propagated.
These observations motivate the use of adaptive temporal fusion, where the balance between current and historical information depends on turbulence severity.

% ===========================================
\section{Proposed Method}

\begin{figure*}[t!]
    \centering
    \includegraphics[width=0.8\linewidth]{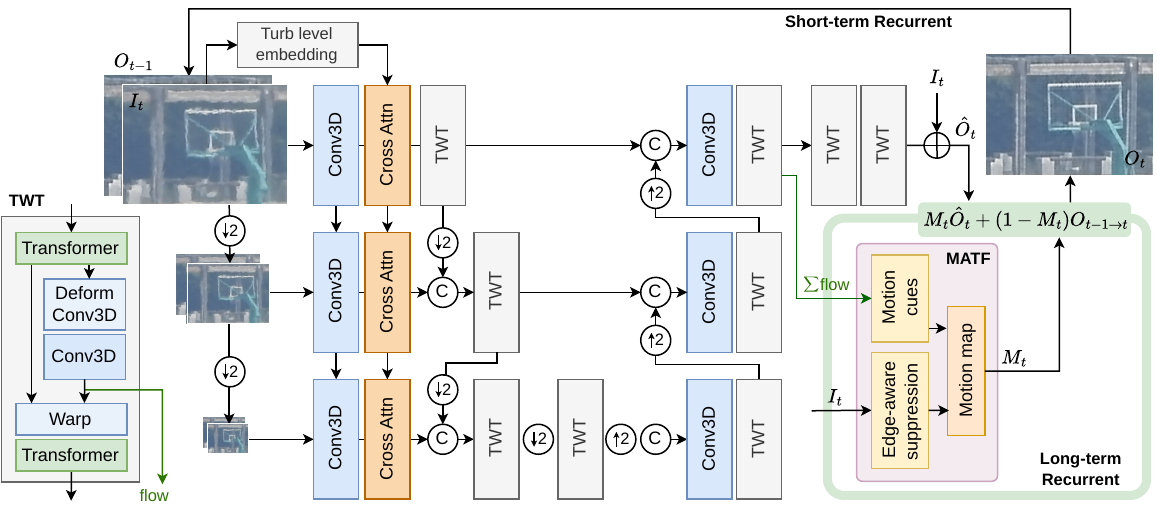}
    \caption{Overview of our proposed turbulence restoration framework.
    }
    \label{fig:architecture}
\end{figure*}

% -------------------------------------------
%\subsection{Overview of framework}
Our turbulence restoration framework uses a recurrent architecture to exploit both short-term temporal cues and long-term priors. As illustrated in Fig.~\ref{fig:architecture}, at time step $t$ the model takes the current distorted frame $I_t$ and the previous restored output $O_{t-1}$, concatenating to form a two-frame embedding that encodes coarse temporal correspondence and prior restoration context.

The embedding is forwarded to the \emph{Mitigator}, which extracts spatio-temporal features, aligns multi-scale structures, and refines high-frequency details using hierarchical 3D convolutions and transformer-based temporal modelling. \textit{Turbulence-level conditioning} modulates the encoder according to distortion severity, producing an intermediate restoration $\hat{O}_t$. The proposed \emph{Motion-Adaptive Temporal Fusion (MATF)} module then adaptively fuses $\hat{O}_t$ with the motion-compensated previous output, balancing temporal smoothness and motion-preserving sharpness to reduce ghosting and flickering.
The final restored frame $O_t$ is recursively propagated to the next time step, enabling memory-efficient long-horizon aggregation while remaining robust to spatially varying atmospheric turbulence.

% -------------------------------------------
\subsection{Turbulence-Level Conditioning}
\label{ssec:turbcondition}
To enable the model to adapt to varying distortion levels, we introduce a per-frame turbulence level adjustment mechanism. First, the distorted input is evaluated for perceptual quality using a CLIP-IQA model, following analysis in Sec.~\ref{ssec:severe}. We evaluate multiple prompt formulations and find that the terms ``quality” and ``noisiness” provide the most reliable correlation with AT severity on ATSyn-Dynamic (see Appendix), enabling more stable adaptation across distortion levels. Since severe turbulence degrades perceptual quality, we define the turbulence level as the normalized reciprocal of the quality score. This scalar serves as a concise indicator of the overall turbulence severity in the current frame.

This scalar is then transformed into a latent embedding via a lightweight multilayer perceptron (MLP) and injected into the backbone network through a cross-attention modulation layer. Specifically, the turbulence embedding serves as the query, while the encoded spatio-temporal features act as both key and value. Through this turbulence-aware attention mechanism, the model dynamically reweights feature responses based on the estimated distortion level. Consequently, the backbone allocates greater correction capacity during intense turbulence (geometric stabilization and multiscale alignment), while prioritizing detail refinement and texture consistency under weaker turbulence. This mechanism enables the model to operate consistently across varying turbulence intensities without requiring separate models or manually designed heuristics.

% -------------------------------------------
\subsection{Mitigator Architecture}
The {Mitigator} is explicitly designed to model turbulence-induced distortions and to establish reliable spatio-temporal correspondences under spatially varying geometric fluctuations. The architecture comprises three principal components: 
(1) multi-scale feature extraction, 
(2) temporal alignment, and 
(3) feature refinement.

\vspace{2mm}
\noindent\textbf{Multi-Scale Feature Extraction.}
AT produces spatially non-uniform distortions that vary across scales, making hierarchical alignment essential. The concatenated inputs $(I_t, O_{t-1})$ are hence downsampled to half and quarter resolutions to construct a three-scale feature hierarchy. At each scale, low-level spatio-temporal representations are extracted using  3D convolutional blocks. These layers capture local geometric deformations, ripple artifacts, and intensity fluctuations induced by AT. Turbulence-level conditioning is injected at this stage to modulate feature responses according to distortion severity. This conditioning mechanism enables the encoder to adapt its feature extraction behaviour across different turbulence regimes, improving robustness to both subtle and severe distortions.

\vspace{2mm}
\noindent\textbf{Temporal Feature Alignment.}
At each scale, we introduce a \textit{Transformer-Warp-Transformer} (TWT) block. The initial transformer enhances feature embeddings and facilitates attention-driven correspondence reasoning, allowing the network to emphasise stable regions while suppressing severely distorted areas.
A lightweight flow estimation branch predicts forward optical flow fields, which guide a flow-warp module equipped with deformable convolutions~\cite{8237351} to compensate for fine-grained geometric distortions. The second transformer further refines the aligned embeddings. This coarse-to-fine, multi-scale design enables the model to handle both large displacements and subtle local fluctuations before proceeding to reconstruction.

\vspace{1mm}
\noindent\textbf{Feature Refinement.}
Aligned features from all scales are fused and processed by a refinement module composed of two additional TWT blocks. This stage reconstructs high-frequency textures and removes residual distortions that remain after alignment. The resulting features are decoded to produce the intermediate restoration $\hat{O}_t$, subsequently utilised by MATF module for temporally adaptive fusion.

The integration of 3D convolutions, transformer-based correspondence modelling, hierarchical alignment, and flow-guided warping equips the Mitigator to effectively address the stochastic, spatially varying characteristics of AT.

% -------------------------------------------

\subsection{Motion-Adaptive Temporal Fusion (MATF)}
\label{ssec:matf}
%\vspace{-7em}
In the recurrent frameworks, such as RMFAT \cite{Liu2025RMFAT}, each frame is restored by referencing both the current turbulent input and the restored result from the previous time step. While this is effective in stabilizing reconstruction, we observe two common failure cases: (1) if the model relies too heavily on the previous output, fast-moving content tends to produce ghosting or trailing artifacts; (2) if it relies too much on the current input, the image becomes sharp yet unstable, producing noticeable flickering. This is confirmed by our study presented in Appendix~\ref{sec:A_tempfusion}. To address this trade-off, we introduce a lightweight MATF strategy.

MATF performs pixel-wise estimation to determine the relative contributions of the previous restored frame \(O_{t-1}\) and the current network output \(\hat{O}_{t}\). We use the predicted flow (one of the TWT outputs) at the finest level to align the previous restored frame with the current frame. Let \(\mathbf{F}^{0}_{t-1\rightarrow t}\) denote the first flow level from the flow pyramid, consisting of a two-channel flow field representing horizontal and vertical motion. The flow is then resized to the spatial resolution of the current frame:
$\mathbf{F}_{t-1\rightarrow t}
=
\operatorname{Up}\left(\mathbf{F}^{0}_{t-1\rightarrow t}; H,W\right)$,
where \(\operatorname{Up}(\cdot)\) denotes bilinear interpolation.
The previous restored frame \(O_{t-1}\) is then warped to the current frame:
$\bar{O}_{t-1\rightarrow t}
=
\mathcal{W}\left(O_{t-1}, \mathbf{F}_{t-1\rightarrow t}\right)$,
where \(\mathcal{W}(\cdot)\) is differentiable bilinear warping. 

The MATF module computes a motion reliability map from four cues: flow magnitude, photometric discrepancy, forward--backward flow consistency, and image edges. Specifically, the normalized flow magnitude is defined as
\begin{equation}
F_t
=
\mathcal{N}_{p}
\left(
\sqrt{
\left(\mathbf{F}^{x}_{t-1\rightarrow t}\right)^2
+
\left(\mathbf{F}^{y}_{t-1\rightarrow t}\right)^2
}
\right),
\end{equation}
where \(\mathcal{N}_{p}(\cdot)\) denotes percentile-based clipping and normalization. The photometric discrepancy between the current input and the warped previous restoration is computed as
\begin{equation}
E_t
=
\mathcal{N}_{p}
\left(
\frac{1}{C}
\sum_{c=1}^{C}
\left|
I_t^{c}
-
\bar{O}_{t-1\rightarrow t}^{c}
\right|
\right), C=3.
\end{equation}
When backward flow \(\mathbf{F}_{t\rightarrow t-1}\) is available, we also compute a normalized forward--backward consistency error:
\begin{equation}
C_t
=
\mathcal{N}_{p}
\left(
\left\|
\mathcal{W}
\left(
\mathbf{F}_{t\rightarrow t-1},
\mathbf{F}_{t-1\rightarrow t}
\right)
+
\mathbf{F}_{t-1\rightarrow t}
\right\|_2
\right).
\end{equation}
If backward flow is unavailable, we set \(C_t=0\). In addition, we compute an edge map from the current frame using a Sobel operator: $G_t = \operatorname{Sobel}(I_t)$.

The motion uncertainty score is then obtained by combining these cues:
$m_t = aF_t + bE_t + cC_t + dG_t$,
where \(a,b,c,d\) control the relative contributions of the motion, photometric, consistency, and edge cues. Here, we set $a=0.5, b=0.5, c=0.1, d=0.3$.  The score is mapped to \([0,1]\) using a scaled hyperbolic tangent:
$M'_t = \frac{1}{2}
\left(
\tanh(2m_t) + 1
\right)$.

To improve temporal stability of the fusion, we smooth the motion map using an exponential moving average:
\begin{equation}
\tilde{M}_t
=
\begin{cases}
M'_t, & t=1,\\
(1-\beta)M_t + \beta \tilde{M}_{t-1}, & t>1,
\end{cases}
\end{equation}
where \(\beta\) is the Exponential Moving Average (EMA) decay factor (\(\beta=0.8\)). The final temporal fusion map is defined as
$M_t = \operatorname{clip}
\left(
\tilde{M}_t, 0, 1
\right)$.
Thus, low-motion and photometrically consistent regions obtain smaller \(M_t\), allowing stronger reuse of the warped previous restoration, while regions with large motion or poor alignment rely more on the current prediction.

Finally, the scalar fusion map is expanded across color channels and used to blend the warped historical restoration with the current-frame prediction \(\hat{O}_t\):
\begin{equation}
O_t
=
M_t\odot \hat{O}_t  + 
(1-M_t) \odot \bar{O}_{t-1\rightarrow t}.
\end{equation}
Static pixels place greater confidence in the warped previous output to maintain temporal consistency and reduce ripple artifacts, whereas dynamic pixels place greater trust in the current restoration to avoid ghosting and preserve object boundaries. As a result, the final reconstructed frame preserves dynamic details while maintaining global temporal coherence.

\begin{figure*}[t]
    \centering
    \includegraphics[width=1.0\linewidth]{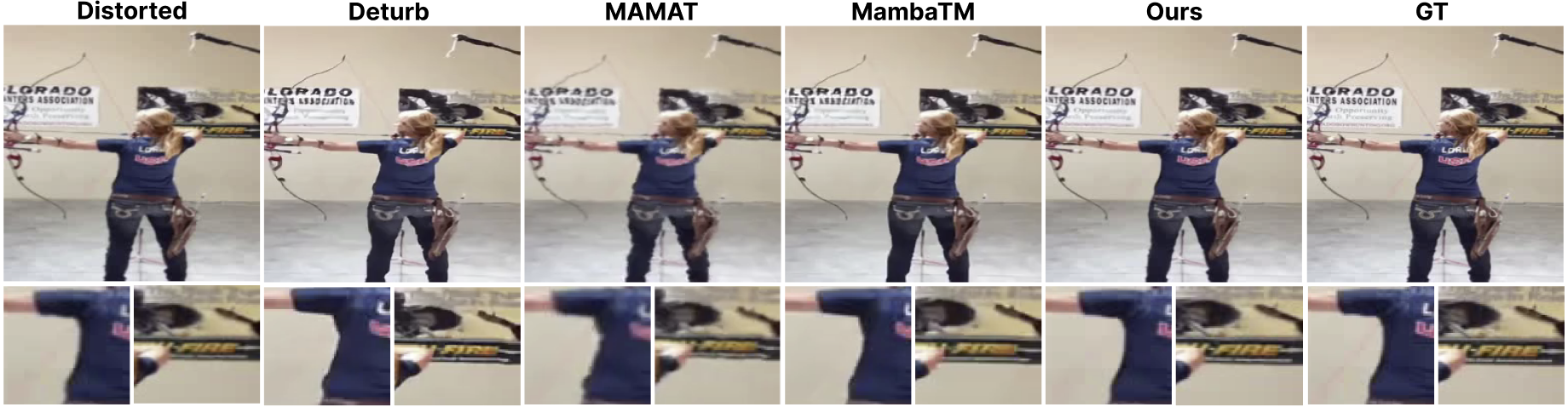}
    \centering
    \caption{Qualitative comparisons on synthetic ATSyn-dynamic dataset.}
    \label{fig:synthetic}
\end{figure*}
% -------------------------------------------
\subsection{Losses}
\label{loss}
Our training objective comprises several complementary losses that enhance spatial fidelity, texture sharpness, and temporal coherence in restored video sequences. Let $T_t$ denote the ground truth frame at time step $t$.

\textit{Reconstruction Loss.}
We adopt the Charbonnier penalty to provide robust pixel-wise supervision under heavy turbulence distortions:
%\begin{equation}
$\mathcal{L}_{\text{rec}}
= \sqrt{(O_t - T_t)^2 + 10^{-6}}$.
%\label{eq:charbonnier}
%\end{equation}

\textit{Wavelet-Domain Loss.}
To better preserve high-frequency structures, we impose a single-level Haar wavelet $\mathcal{H}$ constraint:
%\begin{equation}
$\mathcal{L}_{\text{dwt}}
=
\left\|
\mathcal{H}(O_t) - \mathcal{H}(T_t)
\right\|_{1}$.
%\label{eq:dwtloss}
%\end{equation}

\textit{Laplacian Structural Loss.}
To further enhance edge sharpness and structural fidelity, we penalise discrepancies in the Laplacian responses:
%\begin{equation}
$\mathcal{L}_{\text{lap}}
=
\left\|
\Delta O_t - \Delta T_t
\right\|_{1}$.
%\label{eq:laploss}
%\end{equation}

\textit{Static-Aware Temporal Consistency Loss.}
The MATF module produces a spatial mask $S_t=1-M_t$ that identifies static regions.  
To maintain temporal coherence while avoiding motion-induced artifacts, we enforce consistency between the current output and the warped previous prediction in static areas:
%\begin{equation}
$\mathcal{L}_{\text{temp}}
=
\left\|
S_t \odot O_t
-
S_t \odot \mathcal{W}(O_{t-1})
\right\|_{1}$.
%\label{eq:temploss}
%\end{equation}

\textit{Multi-Step Flow Consistency Loss.}
Beyond the adjacent frame, we additionally encourage coherence with multiple historical predictions ($K=3$) using the flow fields estimated at different decoding stages.  
For each offset $k$, the earlier output $O_{t-k}$ is warped using the corresponding multi-scale flow $F_{t}^{(k)}$.
%\begin{equation}
%$\widetilde{O}_{t}^{(k)}
%=
%W\!\left(O_{t-k}, F_{t}^{(k)}\right)$.
%\end{equation}
Then, $\mathcal{L}_{\text{flow}}=\sum_{k=1}^K \frac{1}{2^k}||O_t - \mathcal{W}\!\left(O_{t-k}, F_{t}^{(k)}\right) ||_1$.

\textit{Overall Objective.}
The final training objective integrates all components:
\begin{equation}
\mathcal{L}
=
\mathcal{L}_{\text{rec}}
+ \lambda_{\text{dwt}} \mathcal{L}_{\text{dwt}}
+ \lambda_{\text{lap}} \mathcal{L}_{\text{lap}}
+ \lambda_{\text{temp}} \mathcal{L}_{\text{temp}}
+ \lambda_{\text{flow}} \mathcal{L}_{\text{flow}}.
\label{eq:final_obj}
\end{equation}
We set $\lambda_{\text{dwt}}$, $\lambda_{\text{lap}}$, $\lambda_{\text{temp}}$, and $\lambda_{\text{flow}}$ to 0.01, 0.15, 0.05, and 0.01, respectively. The loss terms collectively guide the model to reconstruct sharp details, stabilize temporal transitions, and remain robust across diverse turbulence intensities.

% -----------------------------------------
\begin{table*}[t]
\caption{Performance comparison on the ATSyn-dynamic set. The \colorbox{lightred}{best}, \colorbox{lightorange}{second best}, and \colorbox{lightyellow}{third best} results are highlighted.}
\centering
\resizebox{1\linewidth}{!}{
\begin{tabular}{l|c|c|c|c}
\toprule
\textbf{Turbulence Level} &
\textbf{Weak} &
\textbf{Medium} &
\textbf{Strong} &
\textbf{Overall} \\
\midrule
\textbf{Methods} &
PSNR / SSIM / LPIPS &
PSNR / SSIM / LPIPS &
PSNR / SSIM / LPIPS &
PSNR / SSIM / LPIPS \\
\midrule

RNN-MBP~\cite{Zhu_Dong_Pan_Liang_Huang_Fu_Wang_2022} &
27.9243 / 0.8438 / 0.2096 &
27.4742 / 0.8210 / 0.2178 &
26.0812 / 0.7900 / 0.2511 &
27.2161 / 0.8186 / 0.2245 \\

ESTRNN~\cite{zhong2023real} &
28.9805 / 0.8622 / 0.2005 &
28.3338 / 0.8472 / 0.2063 &
26.8897 / 0.8076 / 0.2480 &
28.1347 / 0.8407 / 0.2169 \\

VRT~\cite{10462902} &
28.8453 / 0.8625 / 0.1831 &
28.2628 / 0.8492 / 0.1865 &
26.7492 / 0.8217 / 0.2207 &
28.0179 / 0.8442 / 0.1954 \\

RVRT~\cite{NEURIPS2022_02687e7b} &
29.8950 / 0.8799 / 0.1806 &
29.1658 / 0.8686 / 0.1855 &
27.6827 / 0.8309 / 0.2221 &
28.9332 / 0.8656 / 0.1957 \\
\midrule

TSRWGAN~\cite{Jin2021TurbulenceNet} &
27.0844 / 0.8435 / 0.2141 &
26.7046 / 0.7915 / 0.2221 &
25.4230 / 0.7358 / 0.2671 &
26.4541 / 0.7927 / 0.2325 \\

TMT~\cite{Zhang2024TMT} &
29.1183 / 0.8654 / 0.1820 &
28.5050 / 0.8524 / 0.1841 &
26.9744 / 0.8110 / 0.2206 &
28.2665 / 0.8430 / 0.1942 \\

DATUM~\cite{Zhang2024DATUM} &
 30.2058 /  0.8867 /  0.1788 &
 29.6203 /  0.8783 /  0.1825 &
 28.2550 /  0.8456 /  0.2188 &
 29.4222 /  0.8714 /  0.1919 \\

MambaTM~\cite{Zhang_2025_MambaTM} &
\cellcolor{lightyellow}30.8736 / \cellcolor{lightyellow}0.8991 / \cellcolor{lightyellow}0.1425 &
\cellcolor{lightyellow}30.0816 / \cellcolor{lightyellow}0.8903 / \cellcolor{lightyellow}0.1426 &
\cellcolor{lightyellow}28.6142 / \cellcolor{lightyellow}0.8601 / \cellcolor{lightyellow}0.1721 &
\cellcolor{lightyellow}29.9151 / \cellcolor{lightyellow}0.8843 / \cellcolor{lightyellow}0.1516 \\

\hline
ReMATF (Ours) (ATSyn-dy only)   & \cellcolor{lightorange}31.1248 / \cellcolor{lightorange}0.9127 / \cellcolor{lightorange}0.0843 & \cellcolor{lightorange}30.6515 / \cellcolor{lightorange}0.9058 / \cellcolor{lightorange}0.0921 & \cellcolor{lightorange}28.9874 / \cellcolor{lightorange}0.8712 / \cellcolor{lightorange}0.1189  & 30.4072 / \cellcolor{lightorange}0.8996 / \cellcolor{lightorange}0.0987  \\

ReMATF (Ours) (ATSyn-dy + Real) &
\cellcolor{lightred}31.3748 / \cellcolor{lightred}0.9145 / \cellcolor{lightred}0.0798 &
\cellcolor{lightred}31.0315 / \cellcolor{lightred}0.9085 / \cellcolor{lightred}0.0856 &
\cellcolor{lightred}29.5038 / \cellcolor{lightred}0.8751 / \cellcolor{lightred}0.1106 &
\cellcolor{lightred}30.6913 / \cellcolor{lightred}0.9005 / \cellcolor{lightred}0.0911 \\
\bottomrule
\end{tabular}
}

\label{tab:synthetic}
\end{table*}

% ===========================================
\section{Datasets and Experiment settings}
\label{sec:datasets}

% -------------------------------------------
\subsection{Synthetic datasets}
We use the following synthetic datasets in our experiments. For the generalization analysis in Sec.~\ref{sec:analysis}, we employ the ATSyn-Dynamic dataset, which provides paired turbulence-degraded and clean videos with three labelled severity levels (Weak, Medium, Strong). We follow the official training and testing splits and use the provided level annotations to define the evaluation subsets. We also use the DeTurb dataset, which is generated using a different simulation pipeline and therefore exhibits a distinct distortion distribution from ATSyn. DeTurb provides paired degraded and clean sequences, enabling full-reference evaluation using PSNR, SSIM, and LPIPS. In Sec.~\ref{sec:analysis}, we train separate models on ATSyn-Dynamic and DeTurb to quantify simulator-specific bias through cross-dataset testing.

% -------------------------------------------
\subsection{Real datasets}
As discussed in Sec.~\ref{ssec:syndiversity}, training on multiple synthetic datasets does not fully capture the characteristics of real AT. To address this limitation, we construct a paired dataset comprising 28 videos from real-world turbulent sequences. However, these sequences lack ground truth, as capturing the same dynamic scene under turbulence-free conditions is infeasible.
To approximate ground truth, we generate pseudo-clean references using the statistical restoration method proposed in \cite{Anantrasirichai:Atmospheric:2013}. This approach produces reliable results only in static scenarios where hundreds of distorted frames can be accumulated. To introduce motion while preserving the pseudo ground truth, we simulate dynamic content by randomly cropping the first restored frame and translating the crop window across subsequent frames. This procedure mimics camera translation while maintaining scene consistency. The direction and speed of crop displacement are randomly varied to generate diverse motion patterns in the training data. Further details are provided in Sec.~\ref{ssec:A_realdata}.

During evaluation, we further assess the generalization capability of our method on real-world turbulence using three datasets: CLEAR \cite{Anantrasirichai:Atmospheric:2013}, RLR-AT \cite{xu2024long}, and ATD \cite{Hill2025}. These datasets contain long-range turbulence footage covering a diverse range of scenes and distortion severities. As clean references are unavailable for real recordings, we employ the traditional image restoration method CLEAR to generate pseudo ground truth for static scenes and CLEAR2~\cite{8451755} for dynamic scenes. Consistent with its intended use, we primarily present qualitative comparisons on representative sequences to complement the quantitative results obtained on synthetic benchmarks.

\begin{table*}[t]

    \caption{Performance and efficiency comparison on the real datasets. Speed is
measured in seconds per frame on $480\times384$ images using an NVIDIA RTX
4090. $E'_{w}=E_w \times 10^{-4}$. The \colorbox{lightred}{best}, \colorbox{lightorange}{second-best}, and
\colorbox{lightyellow}{third-best} results are highlighted.}
\centering
\resizebox{1.00\textwidth}{!}{
\begin{tabular}{l|cc|ccccc|ccccc}
\toprule
\textbf{Method} &
\textbf{Params} &
\textbf{Runtime} &
\multicolumn{5}{c|}{\textbf{Static}} &
\multicolumn{5}{c}{\textbf{Dynamic}} \\
\cmidrule(lr){4-8}
\cmidrule(lr){9-13}
&
(M)$\downarrow$ & (s)$\downarrow$ &
PSNR$\uparrow$ &
SSIM$\uparrow$ &
LPIPS$\downarrow$ &
tLPIPS$\downarrow$ &
$E'_{w}\downarrow$ &
PSNR$\uparrow$ &
SSIM$\uparrow$ &
LPIPS$\downarrow$ &
tLPIPS$\downarrow$ &
$E'_{w}\downarrow$ \\
\midrule

TMT \cite{Zhang2024TMT}
& 26.04 & 0.667
& 18.52 & 0.623 & 0.321 & \cellcolor{lightorange}0.0145 & 58.09
& 23.16 & 0.794 & 0.231 & 0.0124 & \cellcolor{lightorange}26.66 \\

DATUM \cite{Zhang2024DATUM}
& 5.75
& \cellcolor{lightyellow}0.031
& 27.41 & 0.915 & 0.172 & 0.0292 & 66.43
& 24.21 & 0.821 & 0.255 & 0.0390 & 37.24 \\

DeTurb \cite{Zou:deturb:2024}
& 58.79 & 1.240
& 27.84 & 0.857 & 0.175 & 0.0163 & \cellcolor{lightorange}57.17
& 22.41 & 0.797
& \cellcolor{lightorange}0.175
& \cellcolor{lightorange}0.0115
& \cellcolor{lightyellow}27.19 \\

MAMAT \cite{hill2025mamat}
& \cellcolor{lightorange}2.83
& 0.089
& \cellcolor{lightyellow}28.85
& \cellcolor{lightyellow}0.937
& \cellcolor{lightyellow}0.155
& 0.0190
& 60.42
& \cellcolor{lightyellow}24.76
& \cellcolor{lightyellow}0.825
& 0.228
& 0.0148
& 34.61 \\

MambaTM \cite{Zhang_2025_MambaTM}
& 6.90
& \cellcolor{lightorange}0.018
& 28.03 & 0.841 & 0.342 & 0.0168 & \cellcolor{lightyellow}57.30
& 24.42 & 0.708 & 0.295
& \cellcolor{lightyellow}0.0117
& 30.81 \\

RMFAT \cite{Liu2025RMFAT}
& \cellcolor{lightred}2.60
& \cellcolor{lightred}0.008
& \cellcolor{lightorange}28.95
& \cellcolor{lightorange}0.941
& \cellcolor{lightorange}0.148
& \cellcolor{lightyellow}0.0150
& 60.71
& \cellcolor{lightorange}25.01
& \cellcolor{lightorange}0.832
& \cellcolor{lightyellow}0.188
& 0.0123
&  34.06 \\

\midrule

{ReMATF (Ours)}
& \cellcolor{lightorange}{2.70}
& \cellcolor{lightred}{0.008}
& \cellcolor{lightred}{28.97}
& \cellcolor{lightred}{0.947}
& \cellcolor{lightred}{0.142}
& \cellcolor{lightred}{0.0144}
& \cellcolor{lightred}{49.57}
& \cellcolor{lightred}{27.84}
& \cellcolor{lightred}{0.882}
& \cellcolor{lightred}{0.140}
& \cellcolor{lightred}{0.0100}
& \cellcolor{lightred}{22.63} \\

\bottomrule
\end{tabular}
}

\label{tab:comparison_real}
\end{table*}

% -------------------------------------------
\subsection{Experiment settings}
We train our model over 100 epochs. Training sequences are randomly cropped into $256 \times 256$ patches with a batch size of $1$. The optimizer is Adam with an initial learning rate of $5 \times 10^{-5}$, decayed by a StepLR scheduler that halves the rate every 5 epochs.
We maintain a short buffer of past predictions and align them recursively using multi-scale optical flow fields estimated by the model.
%
%In preliminary experiments, we observed that some sequences maintain persistently high reconstruction errors with little improvement during training. Such samples provide limited learning signal and can negatively affect optimization by over-weighting gradients from uninformative examples. To address this, we employ a simple learnability-based filtering strategy that removes unlearnable or redundant samples with unreliable supervision. Specifically, we analyze sequence-level loss trajectories and exclude samples whose reconstruction error shows minimal improvement over time. Applying this procedure to the synthetic dataset retains 878 of the original 4,350 training sequences, improving optimization stability and reducing the influence of severely corrupted samples. After filtering, we incorporate real AT sequences with pseudo ground truth generated using synthetic translational motion, enabling the model to learn more realistic dynamic content.
%
We report frame-averaged PSNR, SSIM, and LPIPS, and select checkpoints based on the best validation PSNR+SSIM score. Comparisons are made against state-of-the-art methods, including TSRWGAN~\cite{Jin2021TurbulenceNet}, 
TMT~\cite{Zhang2024TMT}, 
DATUM~\cite{Zhang2024DATUM}, 
MAMAT~\cite{hill2025mamat}, 
and MambaTM~\cite{Zhang_2025_MambaTM}.
\section{Results and Discussion}

% -------------------------------------------
\subsection{Performance on Synthetic Data}

\noindent
To evaluate performance under controlled turbulence, we conduct experiments on the ATSyn-dynamic dataset. Quantitative results across three turbulence levels, as reported in the MambaTM paper, are shown in Tab.~\ref{tab:synthetic}. ReMATF achieves state-of-the-art performance across all metrics and consistently outperforms previous methods under both weak and strong turbulence, demonstrating strong robustness to varying distortion levels. These results indicate that the proposed cyclic recovery framework effectively leverages temporal information to mitigate turbulence-induced artifacts.
Fig.~\ref{fig:synthetic} presents qualitative comparisons on representative sequences from ATSyn-dynamic. While DeTurb produces relatively clear results, it often suffers from temporal instability and structural inconsistencies across frames. In contrast, ReMATF generates temporally stable reconstructions while preserving fine structural details. Moreover, these improvements are achieved with substantially lower model complexity than DeTurb. Other methods tend to exhibit residual distortions or oversmoothing, whereas ReMATF better preserves structural fidelity and restores details closer to the ground truth.

\begin{figure*}[t]
    \centering
    \includegraphics[width=1.0\linewidth]{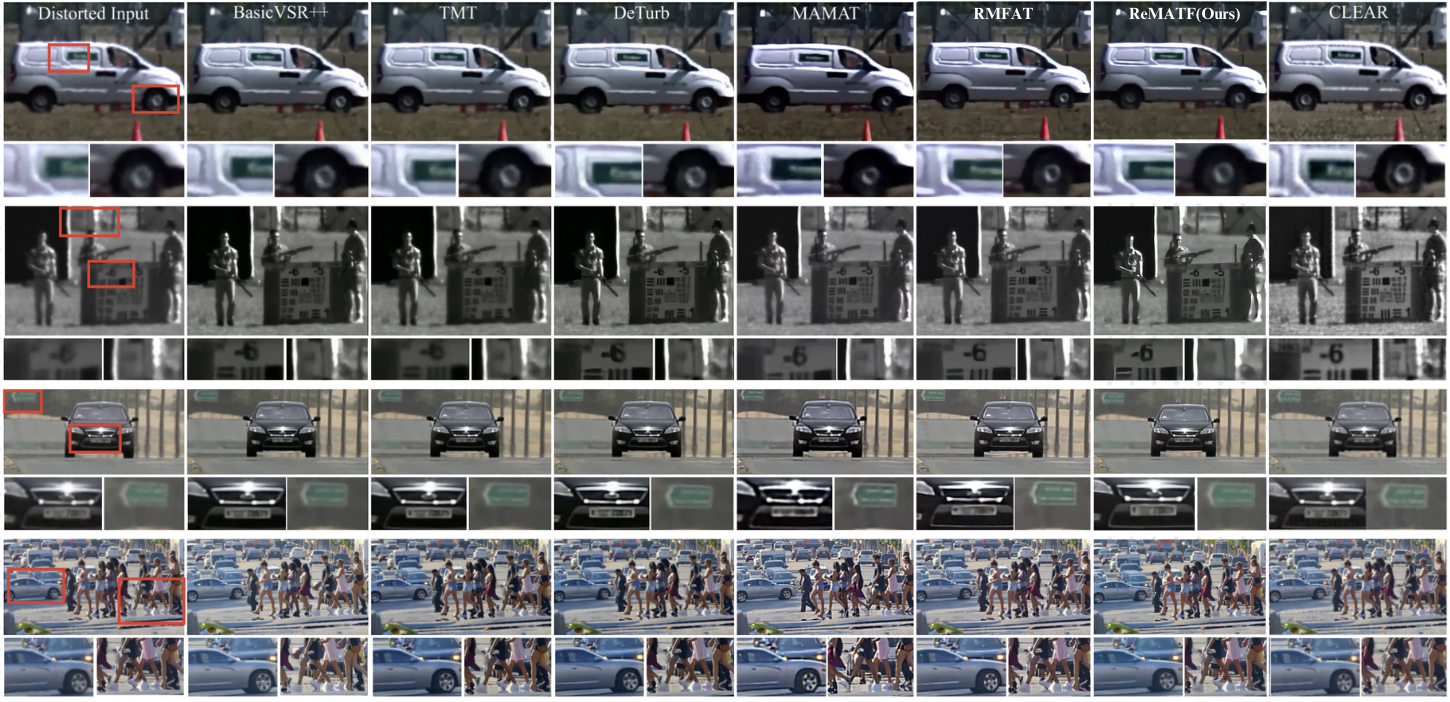}
    \centering
    \caption{Qualitative comparisons on real-world AT from the CLEAR dataset.}
    \label{fig:real}
\vspace{3mm}
    \centering
    \includegraphics[width=1.0\linewidth]{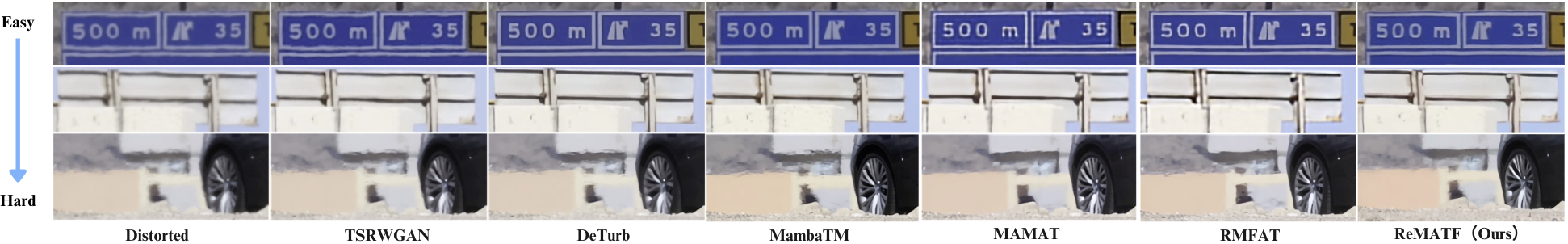}
    \centering
    \caption{Qualitative comparison on real-world ATD \cite{hill_2024_13737763} scenes with increasing turbulence severity (top to bottom).}
    \label{fig:easytohard}
\end{figure*}
% -------------------------------------------
\subsection{Performance on Real Data}
\noindent
%To evaluate the effectiveness of our proposed method in real-world scenarios, we further conducted experiments on a real-world AT dataset. Similar to previous work, we report objective metrics and qualitative comparison results.

\noindent
Tab.~\ref{tab:comparison_real} reports quantitative results on real-world datasets using the publicly available checkpoints of each method. On dynamic scenes, ReMATF outperforms RMFAT by 2.83dB PSNR (11\%), 0.05 SSIM (6\%), and -0.048 LPIPS (25\%). ReMATF achieves the best performance across all evaluation metrics while maintaining a lightweight model and fast inference speed. All methods were trained separately for static and dynamic scenes. As our method primarily targets dynamic content, improvements on static scenes are modest, whereas larger gains are observed on dynamic scenes.

Fig.~\ref{fig:real} shows qualitative comparisons on distorted frames from the CLEAR dataset. ReMATF recovers sharper structures and finer details, particularly in challenging regions such as vehicle boundaries, human silhouettes, and distant objects. In contrast, existing methods either retain noticeable distortions or produce over-smoothed or over-sharpened results, leading to structural degradation.
Fig.~\ref{fig:easytohard} further illustrates restoration results on an ATD scene with progressively increasing turbulence intensity. As distortion severity grows, competing methods often produce unstable structures or blurred textures. In contrast, ReMATF maintains stable reconstructions and stronger structural consistency across turbulence levels, demonstrating improved robustness under severe atmospheric disturbances. Overall compared with previous methods, ReMATF consistently delivers higher reconstruction quality and improved perceptual similarity without increasing computational cost.

\begin{figure}
  \centering
  \includegraphics[width=1\linewidth]{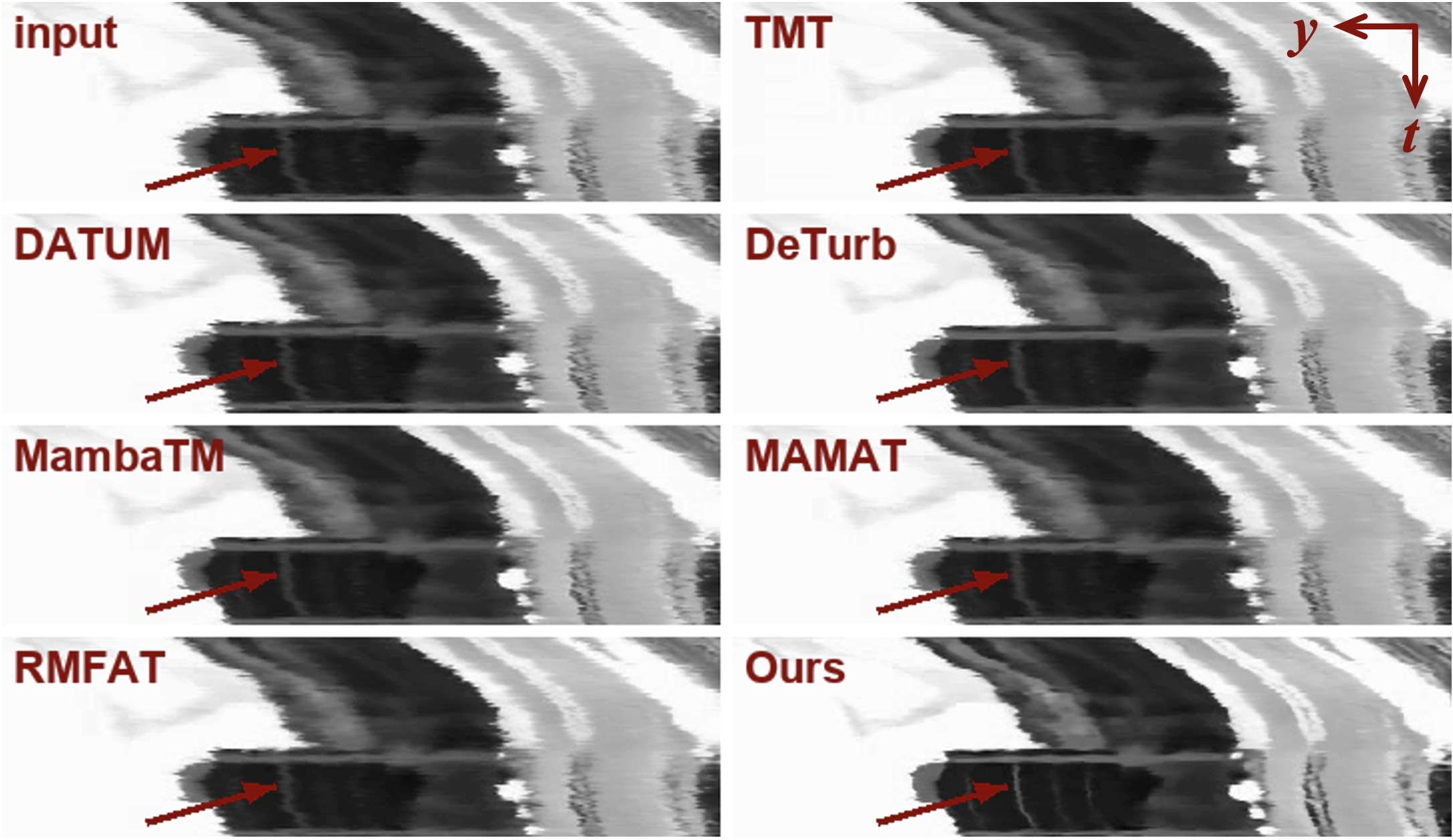}
   \caption{$y$–$t$ plots showing the temporal consistency of each model on the real-world CLEAR dataset. Zoom in for improved visualization of temporal variations and structural continuity.}
   \label{fig:tempconsistent}
\end{figure}

% -------------------------------------------
For {temporal consistency}, we evaluate temporal consistency on dynamic scenes using tLPIPS and warping error ($E_{w}$), with quantitative results reported in Tab.~\ref{tab:comparison_real}. ReMATF achieves the best tLPIPS and $E_{w}$ performance, consistent with the qualitative temporal stability comparisons shown in Fig.~\ref{fig:tempconsistent}. The $y$–$t$ plots demonstrate that ReMATF restores sharper fine details with fewer temporal fluctuations, particularly in the regions highlighted by the red arrows.

% -------------------------------------------
We also evaluate the effect of including real samples in training, which improves PSNR by 2.71 dB and SSIM by 0.048 while reducing LPIPS by 0.041. Additional results are provided in Appendix.

% -------------------------------------------
\subsection{Ablation study}

\begin{table}[t]
\caption{Ablation study on architecture on the \textbf{ATSyn-dynamic} dataset.}
\centering
%\small
\resizebox{0.95\linewidth}{!}{

\begin{tabular}{lcccc}
\toprule
\textbf{Config} & \textbf{PSNR} ↑ & \textbf{SSIM} ↑ & \textbf{LPIPS} ↓ \\
\midrule
w/o MATF & 29.3582 & 0.8810 & 0.1364 \\
w/o multi-scale warp    & 29.9555 & 0.8971 & 0.1013 \\
w/o turb level embedding  & 30.1545 & 0.8879 & 0.1075 \\
\midrule
\textbf{Full (Ours)} & \textbf{30.4072} & \textbf{0.8996} & \textbf{0.0987}\\
\bottomrule
\end{tabular}
}
\label{tab:ablation}
\end{table}

\noindent
To analyze the contributions of each component in the proposed framework, we conducted ablation experiments on the ATSyn-dynamic dataset. The results are shown in Tab.~\ref{tab:ablation}. Removing the proposed MATF module leads to a significant performance degradation, indicating that adaptive fusion between the current recovery result and previous outputs is crucial for maintaining temporal stability and avoiding motion-induced artifacts. Removing the multi-scale distortion module also degrades the recovery quality, demonstrating that accurate motion alignment is essential for handling spatially varying turbulent distortions. Furthermore, removing the turbulence-level embedding also degrades performance, suggesting that explicitly conditioning on turbulence intensity helps the model adjust its recovery behaviour across different distortion levels. When all components are enabled, the complete model achieves the best performance across all evaluation metrics, confirming that each proposed module contributes positively to overall recovery quality.
% ===========================================
\section{Conclusion}
This paper presented ReMATF, a lightweight recurrent framework for mitigating AT in video sequences. ReMATF processes only two frames at a time while maintaining spatial fidelity and temporal consistency. The proposed architecture combines a multi-scale encoder–decoder with temporal warping and a novel MATF module, which performs per-pixel fusion between the warped previous output and the current prediction to stabilize temporal dynamics without increasing memory requirements. We also propose an AT-level estimation module to guide network learning and create a dynamic training set from static real-AT videos.
Experiments on both synthetic and real AT datasets show that ReMATF consistently improves restoration quality while providing significantly faster inference than recent multi-frame transformer-based methods. 

%\clearpage  % TODO FINAL: This \clearpage needs to be removed from both review and camera-ready versions.

{
\small
\bibliographystyle{splncs04}
\bibliography{main}
}

\setcounter{figure}{0}
\setcounter{table}{0}
\renewcommand{\thefigure}{S\arabic{figure}}
\renewcommand{\thetable}{S\arabic{table}}

\renewcommand{\thesection}{S\arabic{section}}
\renewcommand{\thesubsection}{S\arabic{section}.\arabic{subsection}}
\renewcommand{\thesubsubsection}{S\arabic{section}.\arabic{subsection}.\arabic{subsubsection}}

\setcounter{section}{0}

\begin{table*}[tb]
\centering
\caption{Motion-binned analysis of fixed temporal fusion weights on the Strong test split. Results are averaged over the Weak, Medium, and Strong training settings.}
\label{tab:fixed_weight_motion_strong}
\setlength{\tabcolsep}{4pt}
\renewcommand{\arraystretch}{1.0}
\begin{adjustbox}{width=0.8\textwidth,center}
\begin{tabular}{lc|ccc|ccc|ccc}
\toprule
\textbf{Model} & \textbf{Motion} &
\multicolumn{3}{c|}{\textbf{$w=0.50$}} &
\multicolumn{3}{c|}{\textbf{$w=0.75$}} &
\multicolumn{3}{c}{\textbf{$w=0.90$}} \\
\cmidrule(lr){3-5}
\cmidrule(lr){6-8}
\cmidrule(lr){9-11}
& &
PSNR$\uparrow$ & SSIM$\uparrow$ & LPIPS$\downarrow$ &
PSNR$\uparrow$ & SSIM$\uparrow$ & LPIPS$\downarrow$ &
PSNR$\uparrow$ & SSIM$\uparrow$ & LPIPS$\downarrow$ \\
\midrule
MAMAT   & Slow & 24.3858 & 0.7293 & 0.3361 & 25.3967 & 0.7548 & 0.3089 & \textbf{25.8157} & \textbf{0.7634} & \textbf{0.2946} \\
MambaTM & Slow & 24.9081 & 0.7586 & 0.2644 & 26.1791 & 0.7918 & 0.2313 & \textbf{26.8632} & \textbf{0.8058} & \textbf{0.2154} \\
\midrule
MAMAT   & Fast & \textbf{24.3501} & \textbf{0.7551} & \textbf{0.3337} & 22.4278 & 0.7151 & 0.3853 & 20.2574 & 0.6665 & 0.4477 \\
MambaTM & Fast & \textbf{25.0127} & \textbf{0.7857} & \textbf{0.2631} & 22.7781 & 0.7362 & 0.3263 & 20.4118 & 0.6784 & 0.4026 \\
\bottomrule
\end{tabular}
\end{adjustbox}
\end{table*}

% -------------------------------------------
\section{Additional Analysis of Generalization in AT Mitigation }

\subsection{Effect of Motion on Temporal Fusion}
\label{sec:A_tempfusion}
We further analyze how scene motion affects the preferred temporal fusion behaviour under severe atmospheric turbulence. While stronger temporal accumulation can improve temporal stability in relatively stable regions, its effectiveness may change substantially once object motion becomes large.

When object motion is large, the benefit of stronger temporal accumulation diminishes. In dynamic regions, imperfect alignment and occlusion make recurrent propagation more fragile, such that excessive reliance on previous outputs may introduce blur, ghosting, or structural trailing artifacts. This motivates the analysis of motion as an additional factor influencing the preferred temporal fusion behaviour.

To better understand this discrepancy, we conduct this analysis on synthetic data \cite{Zhang_2024_CVPR}, where object motion can be separated more clearly from the micro-motions induced by AT. Object motion speed is estimated using optical flow, while turbulence severity is given by the synthetic generation protocol \cite{Zhang2024DATUM}. We divide the test clips into slow and fast-motion groups based on the average pixel displacement between consecutive frames. The resulting motion-binned evaluation is shown in Tab.~\ref{tab:fixed_weight_motion_strong}.

The motion-binned results reveal a clearer picture. Under strong turbulence, when motion is slow, larger history weights are indeed beneficial: for both MAMAT and MambaTM, $w=0.90$ achieves the best PSNR, SSIM, and LPIPS. This agrees with the intuition from the $y$--$t$ plane visualisation, where stronger temporal accumulation helps stabilise severe turbulence and suppress ripple-like temporal fluctuations. In contrast, under fast motion, the trend reverses: the more balanced setting $w=0.50$ performs best for both models, while larger history weights lead to pronounced degradation. This indicates that once motion becomes substantial, stronger reliance on historical information amplifies misalignment, occlusion, and error propagation.

\section{More Detail on Real Dataset with Pseudo Motion}

Existing simulators model turbulence physics but suffer from domain bias (\cref{sec:analysis}), while real data lacks supervision. We bridge this gap by combining real turbulence with pseudo-clean references and controlled crop-based motion. We vary the motion speeds and directions, exposing the model to diverse temporal dynamics rather than a fixed linear shift. 
Note that the model is not bound to the pseudo-GT distribution, as training uses two complementary sources: synthetic data with true clean ground truth to anchor sharp content, and real paired clips that capture realistic atmospheric turbulence.

% ---------------------------------
\begin{table}[tb]
\centering
\caption{Composition of the final training set. The synthetic subset corresponds to the filtered ATSyn-Dynamic training data used in the main paper.}
\label{tab:train_comp}
\small
\setlength{\tabcolsep}{6pt}
\renewcommand{\arraystretch}{0.7}
\begin{tabular}{l|c|c}
\toprule
\textbf{Training subset} & \textbf{\# Sequences} & \textbf{\# Frames} \\
\midrule
Filtered ATSyn-Dynamic & 878  & 326,672 \\
Real paired clips & 390  & 124,447 \\
\midrule
ATSyn + Real (Final) & 1,268 & 451,119 \\
\bottomrule
\end{tabular}
\end{table}

\subsection{Real Dataset Construction}
\label{ssec:A_realdata}
To complement the synthetic training data, we construct a paired real AT training set from 28 real turbulent scenes from the RLR-AT dataset\cite{xu2024long}. For each scene, we first generate a pseudo-clean reference image using the statistical restoration method in \cite{Anantrasirichai:Atmospheric:2013}, which is effective for static turbulence sequences with sufficient temporal redundancy. Since reliable ground-truth videos are unavailable for real dynamic AT, we create paired dynamic training clips by applying the same spatial crop to (i) the distorted turbulent frames and (ii) the pseudo-clean reference image. This preserves the real AT distortions in the input while providing a temporally stable supervisory target.

All generated clips use a fixed spatial resolution of $256 \times 256$. For each scene, we produce two types of crop trajectories. First, we construct scan-based trajectories by traversing the image in a random serpentine order with a 6-pixel stride. Intermediate crop positions are further densified using 2-pixel steps to obtain smoother motion, and the resulting trajectory is randomly split into 3-5 scan clips of different lengths. Second, we generate 10 additional random clips per scene. For each random clip, the clip length is uniformly sampled from 240 to 480 frames, while the crop velocity is randomly sampled from 4 to 10 pixels per frame with a random direction. When the crop reaches an image boundary, the motion direction is reflected to keep the crop within the valid image region. The same crop coordinates are used for the turbulent sequence and the pseudo-clean reference, getting frame-wise paired supervision under simulated camera motion.

Across the 28 real scenes, this procedure produces 110 scan clips containing 23,274 frames and 280 random clips containing 101,173 frames, resulting in 390 paired real training clips with a total of 124,447 frames. To ensure consistent spatial dimensions within each scene, all turbulent frames are normalised to the dominant frame resolution before crop generation when necessary, which affects 46 frames in total. For reproducibility, we store the crop trajectory of each generated clip, including the start position, motion parameters, and per-frame crop coordinates. The composition of the final mixed training set is summarised in Tab.~\ref{tab:train_comp}.

%-----------------------------
\subsection{Additional Results with ATSyn + Real Training}
We further evaluate the effect of augmenting the filtered ATSyn-Dynamic training set with the paired real clips constructed above. The resulting mixed training set contains 1,268 sequences and 451,119 frames in total (Tab.~\ref{tab:train_comp}). We compare models trained on ATSyn only and ATSyn + Real on both the synthetic ATSyn-Dynamic benchmark and the real dynamic AT benchmark used in the main paper.

As shown in Tab.~\ref{tab:atsyn_real_results}, augmenting the filtered synthetic training set with the proposed paired real clips improves performance on both the synthetic and real dynamic AT benchmarks. On ATSyn-Dynamic, the mixed training strategy increases PSNR from 30.4072 to 30.6913, SSIM from 0.8996 to 0.9005, and reduces LPIPS from 0.0987 to 0.0911, indicating that the inclusion of real paired clips does not compromise synthetic-domain performance. More importantly, the gains on real dynamic AT are substantially larger: PSNR improves from 25.13 to 27.84, SSIM from 0.834 to 0.882, and LPIPS decreases from 0.181 to 0.140. These results suggest that the proposed real paired clips provide complementary supervision beyond synthetic simulation alone, particularly for dynamic real-world turbulence where appearance changes and motion patterns are not fully captured by synthetic distortions. Combining ATSyn-Dynamic with the constructed real paired clips leads to better generalization while preserving strong restoration quality on the original synthetic benchmark.

\begin{table}[tb]
\centering
\caption{Effect of augmenting ATSyn-Dynamic with the proposed real paired clips. We report performance on both the synthetic ATSyn-Dynamic benchmark and the real dynamic AT benchmark.}
\label{tab:atsyn_real_results}
\setlength{\tabcolsep}{4pt}
\renewcommand{\arraystretch}{1.0}
\begin{adjustbox}{width=\columnwidth,center}
\begin{tabular}{l|ccc|ccc}
\toprule
\multirow{2}{*}{\textbf{Training data}} &
\multicolumn{3}{c|}{\textbf{ATSyn-Dynamic (Overall)}} &
\multicolumn{3}{c}{\textbf{Real (Dynamic)}} \\
\cmidrule(lr){2-4}
\cmidrule(lr){5-7}
& PSNR$\uparrow$ & SSIM$\uparrow$ & LPIPS$\downarrow$
& PSNR$\uparrow$ & SSIM$\uparrow$ & LPIPS$\downarrow$ \\
\midrule
ATSyn only   & 30.4072 & 0.8996 & 0.0987 & 25.13 & 0.834 & 0.181 \\
ATSyn + Real & \textbf{30.6913} & \textbf{0.9005} & \textbf{0.0911} & \textbf{27.84} & \textbf{0.882} & \textbf{0.140} \\
\bottomrule
\end{tabular}
\end{adjustbox}
\end{table}

As shown in \cref{tab:atsyn_real_results}, augmenting the filtered synthetic training set with the proposed paired real clips improves performance on both the synthetic and real dynamic AT benchmarks. On ATSyn-Dynamic, the mixed training strategy increases PSNR from 30.4072 to 30.6913, SSIM from 0.8996 to 0.9005, and reduces LPIPS from 0.0987 to 0.0911, indicating that the inclusion of real paired clips does not compromise synthetic-domain performance. More importantly, the gains on real dynamic AT are substantially larger: PSNR improves from 25.13 to 27.84, SSIM from 0.834 to 0.882, and LPIPS decreases from 0.181 to 0.140. These results suggest that the proposed real paired clips provide complementary supervision beyond synthetic simulation alone, particularly for dynamic real-world turbulence where appearance changes and motion patterns are not fully captured by synthetic distortions. Overall, combining ATSyn-Dynamic with the constructed real paired clips leads to better generalisation while preserving strong restoration quality on the original synthetic benchmark.

\begin{table*}[tb]

\caption{CLIP-IQA prompt analysis on 300 distorted clips from the training split of ATSyn + Real (100 per severity level,
$19{,}200$ frames). \textbf{Combined} is $\bar{q}_t=(q_t+n_t)/2$, whose complement defines the turbulence descriptor $\tau_t$ (Eq.~\ref{eq:A_turbulence_score}). The corresponding $\tau$ values are $0.679$, $0.782$ and $0.828$. A reliable prompt must stay monotonic and keep both adjacent-level transitions resolvable: only the combined descriptor does so.}

\centering
\footnotesize
\setlength{\tabcolsep}{4pt}
\renewcommand{\arraystretch}{0.95}
\begin{tabular}{lccccc}
\toprule
\textbf{AT level} & \textbf{Quality} & \textbf{Sharpness} &
\textbf{Contrast} & \textbf{Noisiness}$_{\mathrm{clean}}$ & \textbf{Combined} \\
\midrule
\multicolumn{6}{l}{\textit{(a) Distorted input (mean\,$\pm$\,std over frames)}} \\
Weak   & 0.364\,$\pm$\,0.199 & 0.325\,$\pm$\,0.246 & 0.134\,$\pm$\,0.074 & 0.278\,$\pm$\,0.155 & 0.321\,$\pm$\,0.166 \\
Medium & 0.218\,$\pm$\,0.133 & 0.086\,$\pm$\,0.078 & 0.171\,$\pm$\,0.091 & 0.218\,$\pm$\,0.170 & 0.218\,$\pm$\,0.124 \\
Strong & 0.195\,$\pm$\,0.214 & 0.077\,$\pm$\,0.112 & 0.183\,$\pm$\,0.063 & 0.149\,$\pm$\,0.115 & 0.172\,$\pm$\,0.146 \\
\midrule
\multicolumn{6}{l}{\textit{(b) Ground truth (reference only, not used at inference)}} \\
Weak   & 0.643 & 0.818 & 0.157 & 0.396 & 0.519 \\
Medium & 0.508 & 0.677 & 0.181 & 0.391 & 0.449 \\
Strong & 0.579 & 0.693 & 0.193 & 0.323 & 0.451 \\
\midrule
\multicolumn{6}{l}{\textit{(c) Adjacent-level effect size ($\uparrow$ better; Cohen's $d$)}} \\
Weak\,$\to$\,Medium   & \textbf{0.85} & 1.26 & $-0.45$ & 0.37 & \underline{0.69} \\
Medium\,$\to$\,Strong & 0.12 & 0.10 & $-0.16$ & \textbf{0.48} & \underline{0.34} \\
\bottomrule
\end{tabular}

\label{tab:A_clip_prompt_analysis}
\end{table*}

% ----------------------------------------
\section{Analysis of Text Prompts for Turbulence-Level Conditioning}
\label{sec:A_prompt_analysis}
\noindent
We investigate which CLIP-IQA text prompts provide a reliable proxy for AT severity. Specifically, we evaluate the four built-in prompts ``quality'', ``sharpness'', ``contrast'' and ``noisiness'', which correspond to the antonym pairs (``Good photo'', ``Bad photo''), (``Sharp photo'', ``Blurry photo''), (``High contrast photo'', ``Low contrast photo'') and (``Clean photo.'', ``Noisy photo.''), where the reported score is the softmax probability of the first term. The analysis uses 300 distorted clips drawn from the training split of ATSyn + Real (100 clips from each of the Weak, Medium and Strong subsets), comprising $19{,}200$ frames in total.%, so that no test data is used for prompt selection.

For turbulence-level conditioning, a prompt provides a reliable severity signal only if it varies monotonically with the annotated severity and keeps every adjacent transition resolvable. A high overall correlation is not sufficient if it is dominated by a single easy contrast. Therefore, for each prompt, we report the per-level mean and standard deviation over frames, the corresponding ground-truth scores, and the effect size (Cohen's $d$) of each adjacent-level transition. Ground-truth scores are reported for reference only. At inference, $\tau_t$ is computed from the distorted input alone.

As shown in Tab.~\ref{tab:A_clip_prompt_analysis}, the ``quality'' score drops sharply from Weak to Medium ($0.364\!\to\!0.218$, $d=0.85$) but then nearly saturates ($0.218\!\to\!0.195$, $d=0.12$), so it cannot resolve the two most severe levels. ``Sharpness'' responds most strongly to the onset of blur ($0.325\!\to\!0.086$, $d=1.26$), but it collapses almost completely between Medium and Strong ($0.086$ versus $0.077$, $d=0.10$). Its comparatively high global rank correlation with severity ($\rho=-0.524$ over $19{,}200$ frames) is therefore driven almost entirely by the Weak-versus-rest contrast and does not indicate a reliable estimate at high severity; the prompt effectively behaves as a near-binary blur detector rather than a continuous severity measure.
``Contrast'' moves in the opposite direction ($0.134\!\to\!0.183$), since turbulence-induced intensity fluctuations can increase local contrast. More decisively, its scores on distorted frames are almost identical to those on the corresponding ground truth (Tab.~\ref{tab:A_clip_prompt_analysis}b, gaps $\leq 0.023$), i.e.\ this prompt barely distinguishes turbulent from clean content at all.
``Noisiness'' is the only prompt whose two adjacent transitions are comparable ($d=0.37$ and $d=0.48$), but its Weak-to-Medium response is the weakest among the monotonic prompts.

We therefore combine ``quality'' and ``noisiness'', as their average is the only descriptor that maintains separability between both pairs of adjacent levels, with $d=0.69$ and $d=0.34$, respectively. The two scores are only moderately correlated across frames (Pearson $r=0.595$, Spearman $\rho=0.706$), confirming that they carry non-redundant information rather than duplicating a single quality axis.
Averaging also stabilizes the signal over time: the mean within-clip temporal standard deviation of $\tau_t$ is $0.057$, compared with $0.069$ and $0.072$ when $\tau_t$ is derived from \textit{quality} or \textit{noisiness} alone. Since $\tau_t$ is estimated per frame and injected into a recurrent restoration network, this roughly $20\%$ reduction in frame-to-frame jitter directly benefits temporal
consistency.

Let $q_t$ denote the quality score and $n_t$ the clean-side score returned by the noisiness prompt for frame $t$. Following the main
paper, we denote by $\bar{q}_t = (q_t + n_t)/2$ the combined CLIP-IQA quality score of frame $t$, obtained by averaging the scores of the two selected prompts. The turbulence level is defined to vary inversely with this combined score, i.e.\ as its normalized complement:
\begin{equation}
    \tau_t = \mathrm{clip}_{[0,1]}\!\left(1 - \bar{q}_t\right)
           = \mathrm{clip}_{[0,1]}\!\left(1 - \frac{q_t+n_t}{2}\right),
    \label{eq:A_turbulence_score}
\end{equation}
where a larger $\tau_t$ indicates stronger turbulence. This yields mean $\tau_t$ values of $0.679$, $0.782$ and $0.828$ for the Weak, Medium and Strong subsets, a stable monotonic ordering. We use equal weighting to avoid fitting additional calibration parameters on this analysis subset and to keep the conditioning signal simple and interpretable. As described in the main paper, $\tau_t$ is subsequently embedded by an MLP and injected into the restoration network.

We note that the three severity subsets are not content-matched: the ground-truth quality scores themselves vary across subsets ($0.643$, $0.508$, $0.579$), so part of the observed variation reflects scene content rather than turbulence. The degradation relative to the corresponding ground truth nevertheless remains monotonic ($0.279$, $0.290$, $0.383$ for quality), indicating that the trend is not an artifact of content differences. Since ground truth is unavailable at inference, the absolute score is the quantity that must be well behaved, which is why prompts are selected on the absolute scores.

% ----------------------------------------
\section{Additional Qualitative Results}
\begin{figure*}
  \centering
  \includegraphics[width=\textwidth,height=0.80\textheight,keepaspectratio]
  {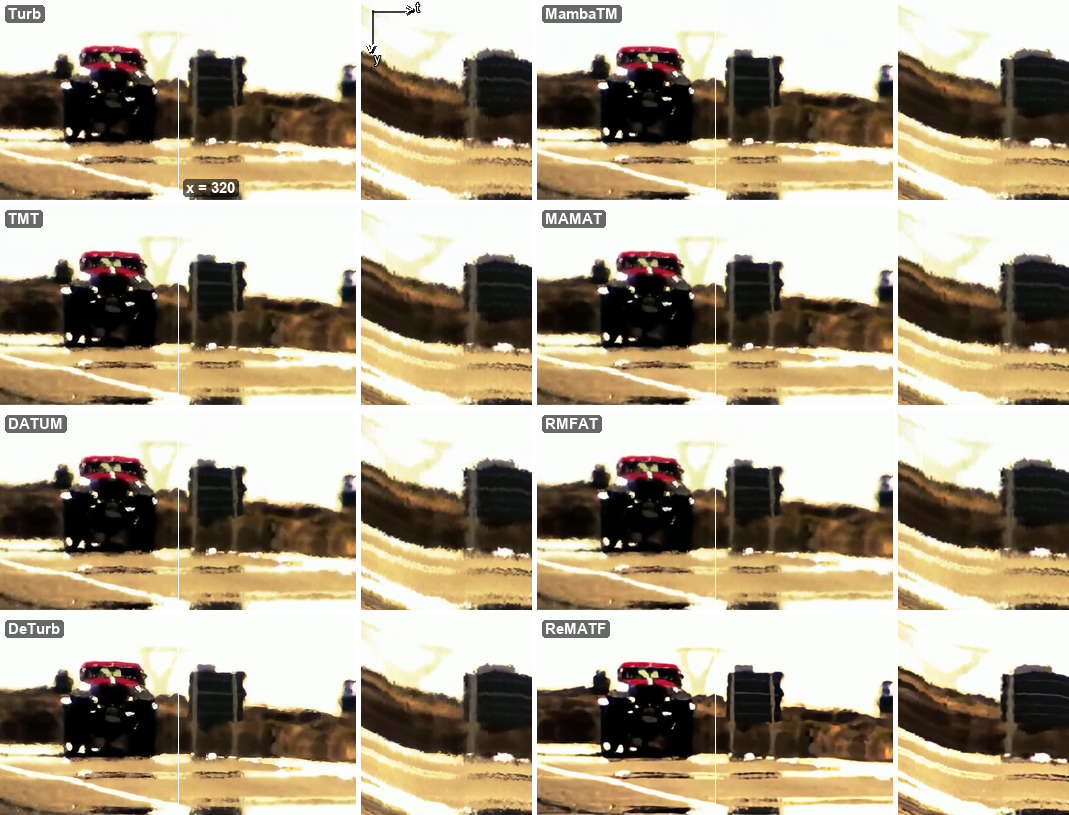}
  \caption{Frame and color $y$--$t$ plane comparison on CLEAR video \textit{Dodge in Heat Wave}.}
  \label{fig:A_clear_yt_01}
\end{figure*}

\begin{figure*}[p]
  \centering
  \includegraphics[width=\textwidth,height=0.80\textheight,keepaspectratio]
  {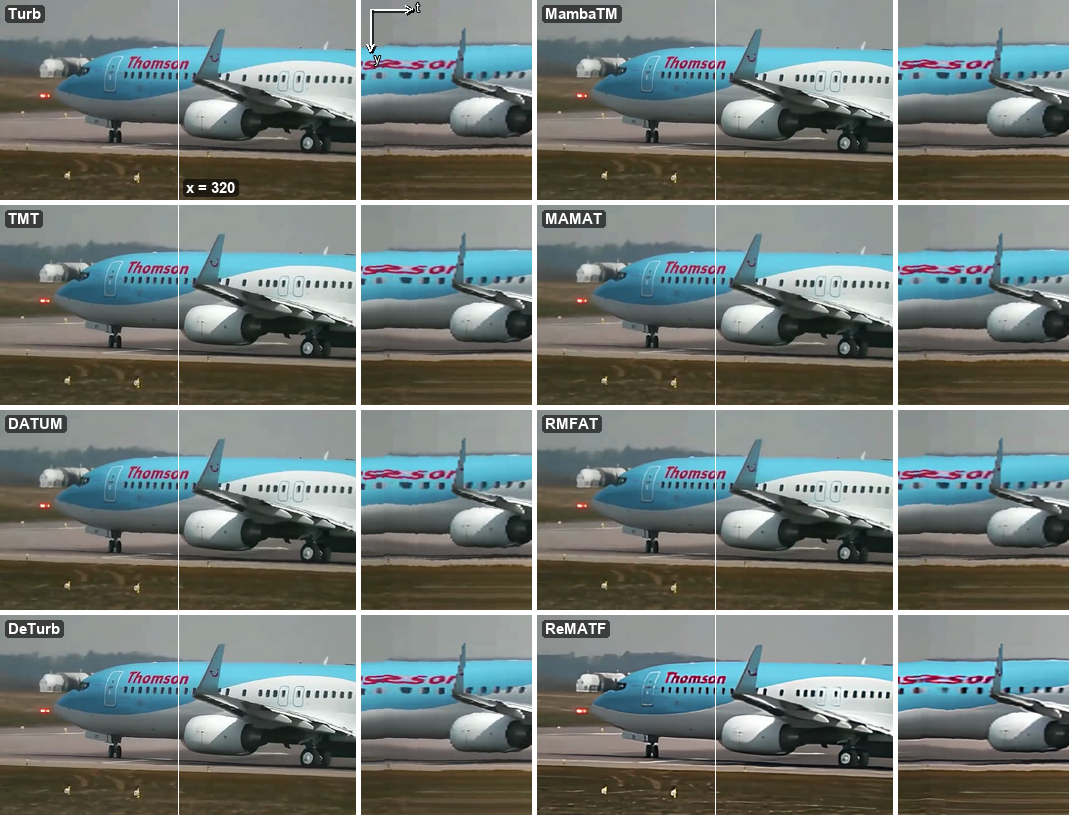}
  \caption{Frame and color $y$--$t$ plane comparison on CLEAR video \textit{East Midlands Airport}.}
  \label{fig:A_clear_yt_02}
\end{figure*}

\begin{figure*}[p]
  \centering
  \includegraphics[width=\textwidth,height=0.80\textheight,keepaspectratio]
  {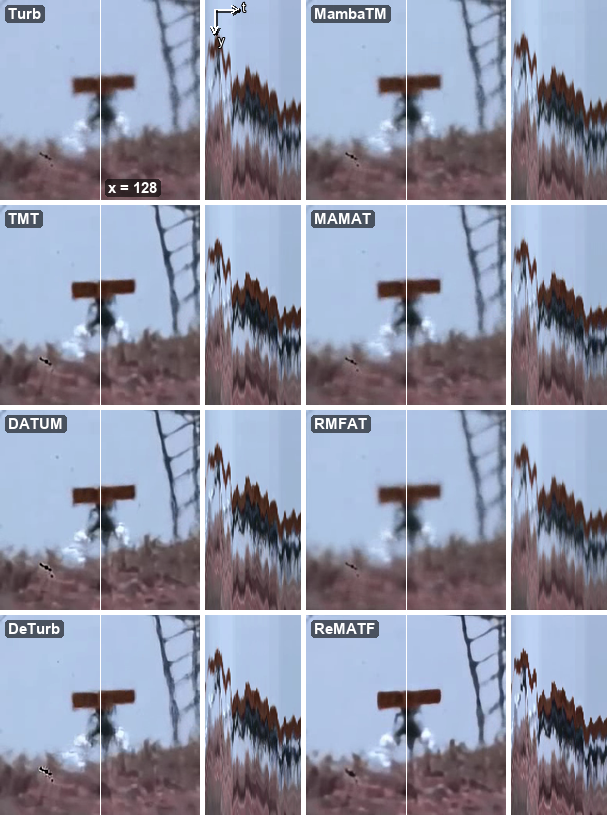}
  \caption{Frame and color $y$--$t$ plane comparison on CLEAR video \textit{Number Plate}.}
  \label{fig:A_clear_yt_04}
\end{figure*}

\begin{figure*}[p]
  \centering
  \includegraphics[width=\textwidth,height=0.80\textheight,keepaspectratio]
  {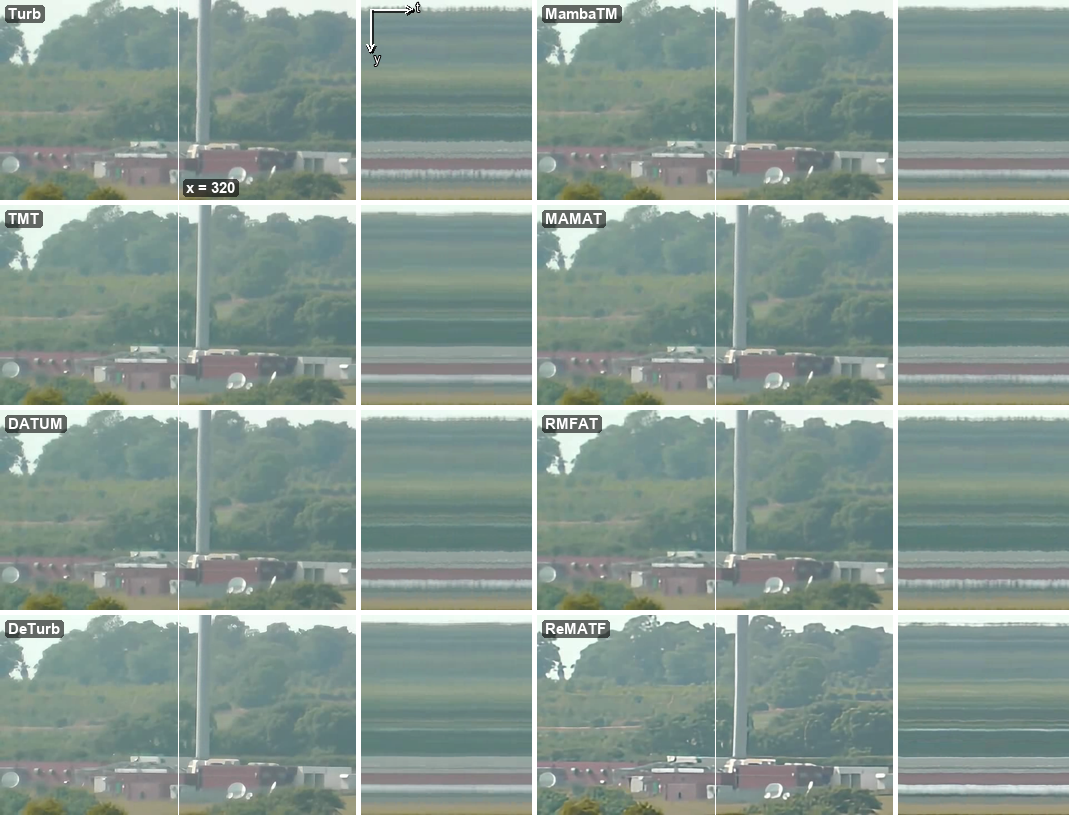}
  \caption{Frame and color $y$--$t$ plane comparison on CLEAR video \textit{Secret Base}.}
  \label{fig:A_clear_yt_05}
\end{figure*}

\begin{figure*}[p]
  \centering
  \includegraphics[width=\textwidth,height=0.80\textheight,keepaspectratio]
  {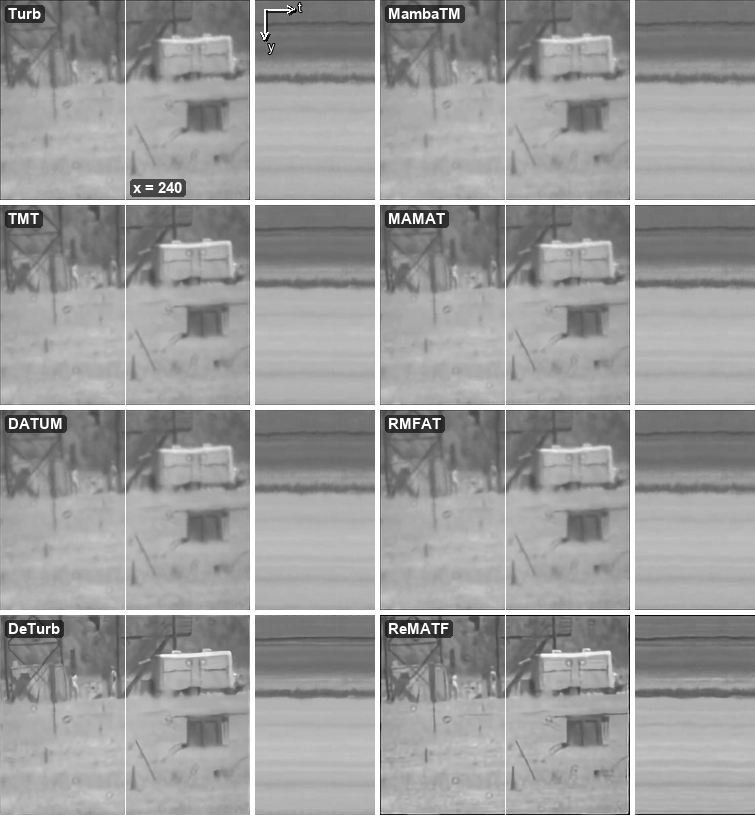}
  \caption{Frame and color $y$--$t$ plane comparison on CLEAR video \textit{Two Men at Distance}.}
  \label{fig:A_clear_yt_06}
\end{figure*}

\noindent
Figures~\ref{fig:A_clear_yt_01}--\ref{fig:A_clear_yt_06}
provide additional qualitative comparisons on six selected CLEAR test videos.
For each method, the left panel shows the restored frame at $t=45$, while the right panel shows the corresponding color $y$--$t$ plane extracted at the marked central column to illustrate temporal consistency. The comparison includes the distorted input (Turb), TMT, DATUM, DeTurb, MambaTM, MAMAT, RMFAT, and ReMATF.

Video \textit{Dodge in Heat Wave} (Figure~\ref{fig:A_clear_yt_01}) demonstrates the strong detail preservation of ReMATF: the vehicle silhouette, small frontal structures, and road boundaries are noticeably sharper. This sharpness is not limited to the selected frame, it remains visible throughout the restored video and is supported by the narrow, coherent structures in the $y$--$t$ plane. In video \textit{East Midlands Airport} (Figure~\ref{fig:A_clear_yt_02}), where the AT distortions are subtle on the aircraft but more severe in the background, ReMATF preserves the main aircraft contours and landing-gear boundaries, as well as the white houses in the background.%In video \textit{Hill House} (Figure~\ref{fig:A_clear_yt_03}), it reconstructs the house structure and roofline particularly well.

The clearest gain is observed in video \textit{Number Plate} (Figure~\ref{fig:A_clear_yt_04}), where ReMATF visibly outperforms all compared methods. It recovers a substantially sharper big yellow plate in the middle and better-defined neighboring utility pole structures on the right, while the corresponding $y$--$t$ plane contains more compact boundaries with reduced temporal waviness. In video \textit{Secret Base} (Figure~\ref{fig:A_clear_yt_05}), the distant scene layers remain stable over time, producing clean and nearly horizontal trajectories with less frame-to-frame fluctuation. In video \textit{Two Men at Distance} (Figure~\ref{fig:A_clear_yt_06}), ReMATF clearly separates the main equipment boundaries. Across the six examples, DeTurb and RMFAT recover comparably sharp edges in some individual frames, whereas TMT, DATUM, MambaTM, and MAMAT tend to produce smoother outputs with less distinct fine structure.

% Display video 0001; source file video_0001.

% Display video 0002; source file video_0003.

% Display video 0003; source file video_0005.
%\begin{figure*}[p]
%  \centering %\includegraphics[width=\textwidth,height=0.80\textheight,keepaspectratio]
 % {yt-plane/video_0005.png}
 % \caption{Frame and color $y$--$t$ plane comparison on CLEAR video \textit{Hill House}.}
 % \label{fig:A_clear_yt_03}
%\end{figure*}

% Display video 0004; source file video_0007.

% Display video 0005; source file video_0009.

% Display video 0006; source file video_0013.

% ----------------------------------------
\section{Failure Cases}
\noindent
Despite its overall spatial and temporal improvements, ReMATF still exhibits several local failure modes. In video \textit{East Midlands Airport} (Figure~\ref{fig:A_clear_yt_02}), most of the aircraft, including the fuselage, engine, and landing gear, remains clear, but the fine `\textit{Thomson}' lettering is slightly blurred. A plausible explanation is that the local motion of this small, high-frequency text is underestimated, assigning it an overly large static-region fusion weight. Because the text moves with the aircraft, recurrent accumulation under imperfect alignment can average slightly misregistered character strokes. This behavior is therefore more
consistent with the region being treated as too static, rather than too dynamic.
~
Video \textit{Two Men at Distance} (Figure~\ref{fig:A_clear_yt_06}) reveals a second limitation. Some artifacts are visible in the trees and other fine background textures of video, even though the main equipment structure remains well defined. Such irregular, high-frequency regions provide ambiguous correspondences under turbulence and possible non-rigid motion, so small alignment errors may be reinforced by recurrent fusion.

\end{document}